\documentclass{article}

\usepackage[preprint]{neurips_2024}

\usepackage[utf8]{inputenc} 
\usepackage[T1]{fontenc}    
\usepackage{hyperref}       
\usepackage{url}            
\usepackage{booktabs}       
\usepackage{amsfonts}       
\usepackage{nicefrac}       
\usepackage{microtype}      
\usepackage{xcolor,colortbl}         
\usepackage{amsmath}
\usepackage{amssymb}
\usepackage{bbm}
\usepackage{pifont}
\usepackage{multirow}
\usepackage[pdftex]{graphicx}
\usepackage{subcaption}
\usepackage{wrapfig}
\usepackage{fancyhdr}
\usepackage{amsthm}
\usepackage{xspace}
\usepackage{makecell}
\usepackage{capt-of}
\usepackage{sidecap}

\newcommand\ie{i.e.\xspace}
\newcommand\eg{e.g.\xspace}

\theoremstyle{plain}

\theoremstyle{definition}

\theoremstyle{remark}

\usepackage[capitalize]{cleveref}
\crefname{section}{Sec.}{Secs.}
\Crefname{section}{Section}{Sections}
\Crefname{table}{Table}{Tables}
\crefname{table}{Tab.}{Tabs.}

\usepackage{amsmath,amsfonts,bm}

\def\1{\bm{1}}

\DeclareMathAlphabet{\mathsfit}{\encodingdefault}{\sfdefault}{m}{sl}
\SetMathAlphabet{\mathsfit}{bold}{\encodingdefault}{\sfdefault}{bx}{n}

\usepackage{graphicx}
\usepackage{amsmath}
\usepackage{amssymb}
\usepackage{booktabs}
\usepackage{threeparttable}
\usepackage{multirow}
\usepackage{xspace}
\usepackage{makecell}
\usepackage{pifont}
\usepackage[dvipsnames]{xcolor}
\usepackage{colortbl}         
\usepackage{standalone}
\usepackage{array}
\usepackage[font=small,labelfont=small]{caption}
\usepackage{pifont}
\usepackage{graphicx}
\usepackage{amsmath}
\usepackage{amssymb}
\usepackage{booktabs}
\usepackage{threeparttable}
\usepackage{makecell}
\usepackage{multirow}
\usepackage{fix-cm}

\usepackage{wrapfig}
\newcommand{\cmark}{\ding{51}}
\newcommand{\xmark}{\ding{55}}

\newcommand{\lidar}{LiDAR\xspace}
\newcommand{\pretrained}{pre-trained\xspace}
\newcommand{\openseed}{OpenSeeD\xspace}
\newcommand{\scalr}{ScaLR\xspace}
\makeatletter
\newcommand\notsotiny{\@setfontsize\notsotiny{6.31415}{7.1828}}
\makeatother

\title{CleverDistiller: Simple and Spatially Consistent Cross-modal Distillation}

\author{%
Hariprasath Govindarajan*$^{2,4}$ \quad Maciej K. Wozniak*$^{1}$ \quad Marvin Klingner$^3$ \\
\textbf{Camille Maurice}$^3$ \quad \textbf{B Ravi Kiran}$^5$ \quad \textbf{Senthil Yogamani}$^6$ \\
$^1$KTH Royal Institute of Technology, Sweden \quad $^2$ Linköping University, Sweden\\
$^3$Qualcomm Arriver Software GmbH \quad $^4$ Qualcomm Auto Ltd Sweden Filial\\
$^5$Qualcomm France, S.A.R.L. \quad $^6$ Qualcomm Technologies, Inc. \\ *equal contribution\\
\texttt{\{hargov,mklingne,cmaurice,ravkira,syogaman\}@qti.qualcomm.com}, \\
\texttt{maciejw@kth.se}\\
}

\begin{document}

\maketitle

\begin{abstract}
Vision foundation models have revolutionized 2D camera-based perception by extracting generalized features for downstream tasks. Recent work applies self-supervised cross-modal knowledge distillation (KD) to transfer these capabilities to 3D LiDAR models, but often relies on complex losses or pseudo-semantic maps.
We introduce CleverDistiller, a self-supervised, cross-modal 2D-to-3D KD framework, introducing simple yet effective design choices. Our method uses a direct feature similarity loss and an MLP projection head to capture complex semantic dependencies without relying on pseudo-semantic maps or explicit semantic supervision. In addition, we improve the 3D spatial reasoning capabilities of learned representations through a self-supervised occupancy prediction task. Experiments on autonomous driving benchmarks show that CleverDistiller achieves state-of-the-art performance in both 3D semantic segmentation and 3D object detection, with up to 10\% mIoU improvement, particularly when fine-tuning with limited data, demonstrating the effectiveness of our approach.
\end{abstract}

\section{Introduction}
\label{sec:intro}

\begin{wrapfigure}{r}{.45\textwidth}
    \centering
    \vspace{-.45cm}
    \includegraphics[width=\linewidth]{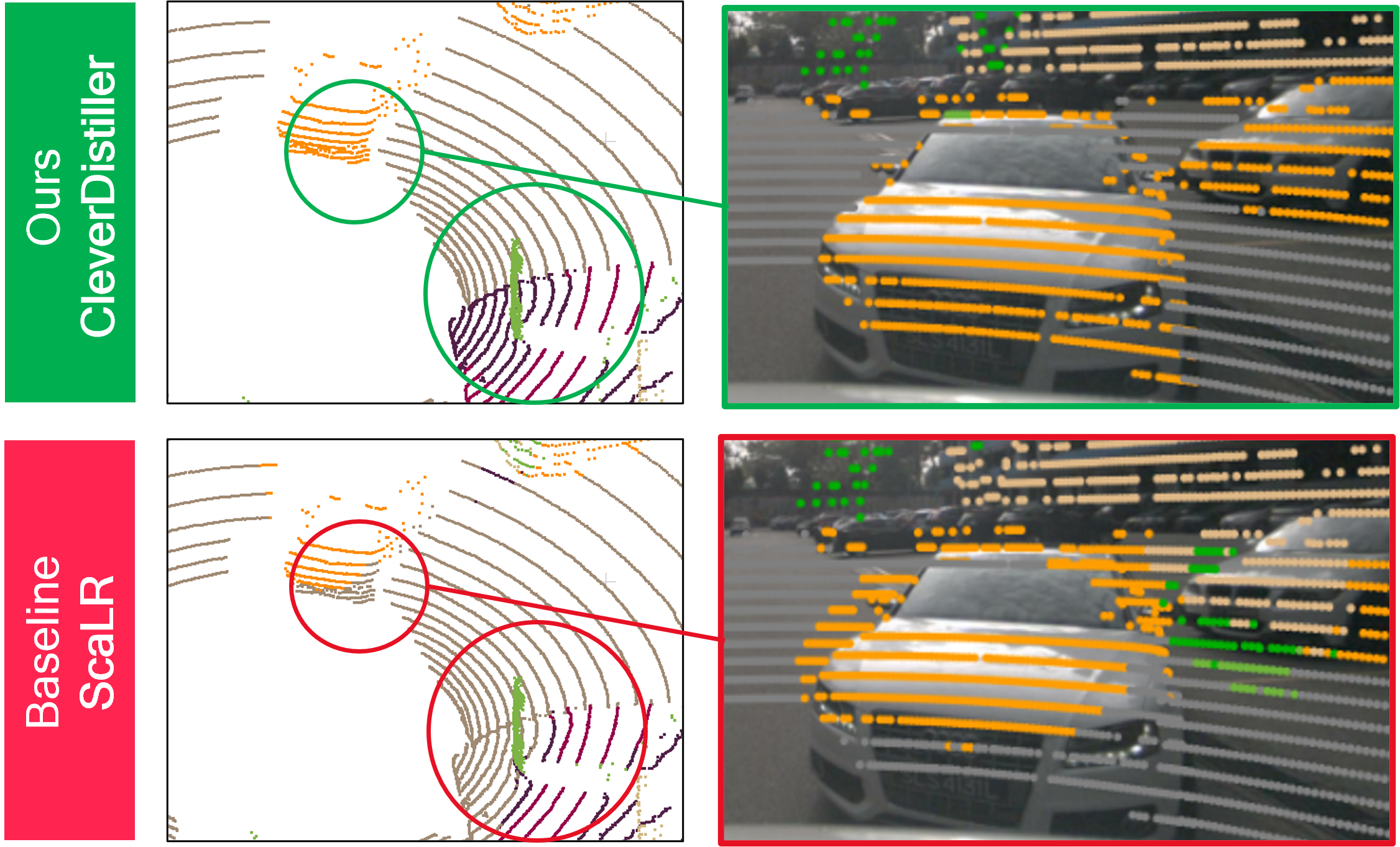}
    \caption{\small We observe how \textbf{CleverDistiller} (top) improves over the baseline \scalr (bottom) by producing spatially consistent semantic outputs.}
    \vspace{-0.3cm}
    \label{fig:front-page}
\end{wrapfigure}For many vision applications such as autonomous driving, improving the generalized understanding of the environment is important to increase safety and autonomy. However, constant updates of sensor types and setups from different manufacturers makes it difficult to train one perception system that excels on all use cases. Most perception models for object detection or semantic segmentation are known for their lack of robustness towards weather conditions, lighting, sensor setup, or driving environments that usually deviate between training and inference~\cite{wozniak2023toward}. Especially, LiDAR-based 3D models are known for their adaptation challenges~\cite{yang2022st3d++,wozniak2024uada3d,tsai2023ms3d++}.
\par

Due to these inherent challenges, recent focus has shifted towards using Vision Foundation Models (VFMs) such as DINO~\cite{sam, dino, dinov2}. These models provide features that are robust and generalize well under domain shift, while also supporting multiple downstream tasks. Although many works address and extend VFMs in camera-based setups~\cite{s3pt,cribo,croc}, our work aims to transfer VFM knowledge to more challenging LiDAR-based 3D networks. Cross-modal knowledge distillation has been used to transfer learned feature representations between the backbones of different sensor modalities. Initial methods focused mainly on distilling LiDAR-based features into camera-based models~\cite{chen2022bevdistill, x3kd}. However, with the emergence of VFMs the reverse direction, to distill generalized semantic 2D features into 3D LiDAR-based networks, has become more relevant~\cite{scalr,seal,superflow,xu2025limoe}. 

Many previous approaches for cross-modal KD have tried to explicitly model the semantic dependencies by using semantic priors, that depend on extra models or preprocessing steps to extract semantically coherent image regions (SAM or OpenSeeD~\cite{sam,openseed,olivine,superflow,xu2025limoe}). They show improvements in performance and robustness, but tend to complicate the cross-modal KD process. On the other hand, ScaLR~\cite{scalr} shows the effectiveness of a simple and scalable distillation solution, but does not achieve SOTA performance. As demonstrated in Fig.~\ref{fig:front-page} \scalr yields semantic outputs that are sometimes inconsistent with the scene's spatial structure. Overall, previous methods are either complicated and involve semantic priors to beat SOTA or are simple and promising but failing to improve over SOTA performance. 

\par
To address these shortcomings, we propose \textbf{CleverDistiller}, a fully self-supervised KD framework that transfers features from a camera-based VFM to a LiDAR-based network. Our approach focuses on obtaining multi-task features by integrating cross-modal KD from a VFM and an auxiliary spatial task, \ie, occupancy prediction~\cite{also}, to capture both semantic and spatial information in a unified framework, thereby improving spatial and semantic feature consistency. Our contributions can be summarized as follows:
\begin{itemize}
    \item We identify linear projection of 3D LiDAR features into 2D space of camera-based VFM features to be a key limitation of cross-modal KD frameworks (overlooked by \textit{all} previous methods). We show that an MLP projection head on top of the 3D LiDAR backbone enables the backbone to learn more informative features that excel across multiple tasks, while also retaining the simplicity of the distillation~(\textbf{\cref{sec:fixing_head,sec:crossmodal}}).
    \item VFM features are semantically rich but lack spatial and geometric information. We incorporate a self-supervised auxiliary task of occupancy prediction to encourage the LiDAR model to additionally encode spatial information. Thus, the general semantic features distilled from the VFM are further enriched to be spatially consistent across the 3D scene~(\textbf{\cref{sec:occupancy}}).
    \item We demonstrate that CleverDistiller outperforms previous approaches in standard 2D-to-3D KD benchmarks for both 3D semantic segmentation and 3DOD. Our model is particularly successful when fine-tuned on limited data, highlighting the generalization capabilities of our learned 3D features (\textbf{\cref{sec:mainresults}}).
\end{itemize}

\section{Related work}
\par
\textbf{Pre-training Vision Models:}
Supervised pre-training on large labeled datasets \cite{dataset_imagenet, imagenet_classification, resnet, deit, deit_3} has being gradually replaced by self-supervised pre-training \cite{ssl_transfer}. It involves training a model using a pretext task on large unlabeled datasets. Currently, self-supervised learning (SSL) methods predominantly use a contrastive \cite{ssl_simclr} or reconstruction formulation~\cite{ibot,dinov2}. Initially, a contrastive objective was considered over data instances in a batch \cite{ssl_instdisc, ssl_simclr} or a memory bank \cite{moco, moco_v2, moco_v3}. Recent works such as DINO \cite{dino} contrast over prototypical clusters \cite{ssl_pcl, deep_cluster, sela, swav, dino_vmf} and learn representations that are highly data efficient \cite{msn, we_ssl}. Reconstruction based methods \cite{mae, beit} typically lack data efficiency \cite{reconst_perception}. Then, DINOv2 \cite{dinov2} consolidated a range of strategies such as masked image modeling \cite{ibot} and produced a model that excels on a wide range of downstream tasks, adopted in many recent works \cite{dinov2_fsod, dinov2_roma, dinov2_tracker}. While pre-training models on the target domain is a promising direction \cite{cribo, s3pt}, there is a lack of off-the-shelf models for autonomous driving, thus in this work we chose DINOv2 as a teacher model to train a 3D student backbone.
\par
\textbf{Pre-training 3D Backbones}
Self-supervised pre-training of 3D backbones started with single object scans (e.g. ShapeNet \cite{shapenet, wu20153d}) similar to ImageNet in the case of images \cite{ssl3d_selfcorr, ssl3d_orientation, ssl3d_pc, ssl3d_pointbert, ssl3d_reconst}. Several methods have been developed to pre-train on complex 3D indoor and outdoor scenes \cite{pointcontrast, ssl3d_proposalcontrast, ssl3d_maskedscene, ssl3d_segcontrast}. Inspired by their vision counterpart, these methods also mainly use contrastive and reconstruction objectives. Some works leverage \lidar specific augmentations \cite{lasermixpp} or use specialized tasks such as occupancy prediction consistent under sparse point sampling \cite{also, agro2024uno} and different mixing strategies for pointclouds \cite{mix3d, lasermix, polarmix}. 
However, these methods are limited by the volume of diverse 3D pre-training data, owing to the difficulty in collecting such a large-scale dataset.
\par 
\textbf{Cross-modal 2D-to-3D Knowledge Distillation:} 
Recently, distilling knowledge from a pre-trained 2D image backbone into a 3D LiDAR backbone has been shown to be an effective pre-training strategy. Initial works focused on single object pointclouds \cite{teacher_autoencoders, zhang2023learning} but recent works have demonstrated that this strategy is also suitable for complex scenes \cite{ppkt, seal}. SLidR~\cite{slidr} groups points and pixels into semantically coherent superpixels/superpoints based on the teacher vision model features. Other works use additional vision models such as SAM~\cite{sam} or \openseed~\cite{openseed} to extract pseudo-semantic regions for the same purpose \cite{seal, superflow}. Recently, some works have also used spatial and temporal consistency to regularize learned representations \cite{seal, superflow, xu2025limoe, olivine, zhang2024contrastive}. Others demonstrated the potential of scaling up different distillation components such as the dataset, the 3D backbone and the 2D backbone while using a simple distillation loss that matches the corresponding point-pixel features directly~\cite{scalr}. However, direct feature comparison impairs the 3D network's ability to learn geometrically rich features. In this work, we propose an MLP projection head and the auxiliary spatial task to guide the 3D network towards learning more spatially aware features.

\section{Method Description}

\subsection{Problem Definition and Motivation}

Given a point cloud $\mathcal{P} = \{ \mathbf{p}_i \mid i = 1, \ldots, N \}$ with $N$ points captured by a LiDAR sensor, where $\mathbf{p}_i \in \mathbb{R}^{4}$ denotes the coordinates of the point and the laser intensity. Let $\mathcal{I}_t = \{ \mathbf{I}_k \mid k = 1, \ldots, M \}$ denote $M$ surround camera images where $\mathbf{I}_k \in \mathbb{R}^{H \times W \times 3}$ denotes an image with height $H$ and width $W$. The objective of cross-modal distillation is to take adventage of a pre-trained vision foundation model $T_{\phi}(\cdot)$ to learn a LiDAR foundation model, $S_{\theta_p}(\cdot)$ by distilling the features from the image modality to the LiDAR modality. 

Many recent works generate a set of class-agnostic or pseudo-semantic superpixels (\eg using text prompts in vision language models) for each image using additional supervised or unsupervised segmentation methods such as SAM \cite{sam} or \openseed \cite{openseed}. 
These superpixels are not ground truths and require careful hyperparameter tuning to obtain suitable segments for a particular domain. Moreover with larger dataset, generating superpixels requires incrementally more time and computational resources.
This complicates the overall distillation process and \textit{we question the need for semantic priors} in the presence of good features from a VFM. Further, most prior works use a contrastive loss~\cite{slidr,seal,olivine,superflow,zhang2024contrastive,xu2025limoe} which creates an unnecessary \textit{self-conflict} issue: contrasting image segments belonging to similar objects. 

In our approach, we find corresponding point and image features, without relying on any additional processing steps or semantic priors. Assuming that the point cloud $\mathcal{P}_t$ and images $\mathcal{I}_t$ are calibrated, we find corresponding points and pixels by projecting the point cloud $\mathbf{p}_i = (x_i, y_i, z_i)$ onto the image plane $(u_i, v_i)$ using the following sensor calibration parameters, $\begin{bmatrix} u_i & v_i & 1 \end{bmatrix}^T = \rho(i) = \frac{1}{z_i} \times \Gamma_K \times \Gamma_{c \leftarrow l} \times \begin{bmatrix} x_i & y_i & z_i \end{bmatrix}^T$,
where $\Gamma_K$ denotes the camera-intrinsic matrix and $\Gamma_{c \leftarrow l}$ is the transformation matrix from LiDAR sensors to surround-view cameras. This enables a direct mapping between points and pixels in the image.
\begin{figure*}[t!]
    \centering
    \includegraphics[width=\linewidth]{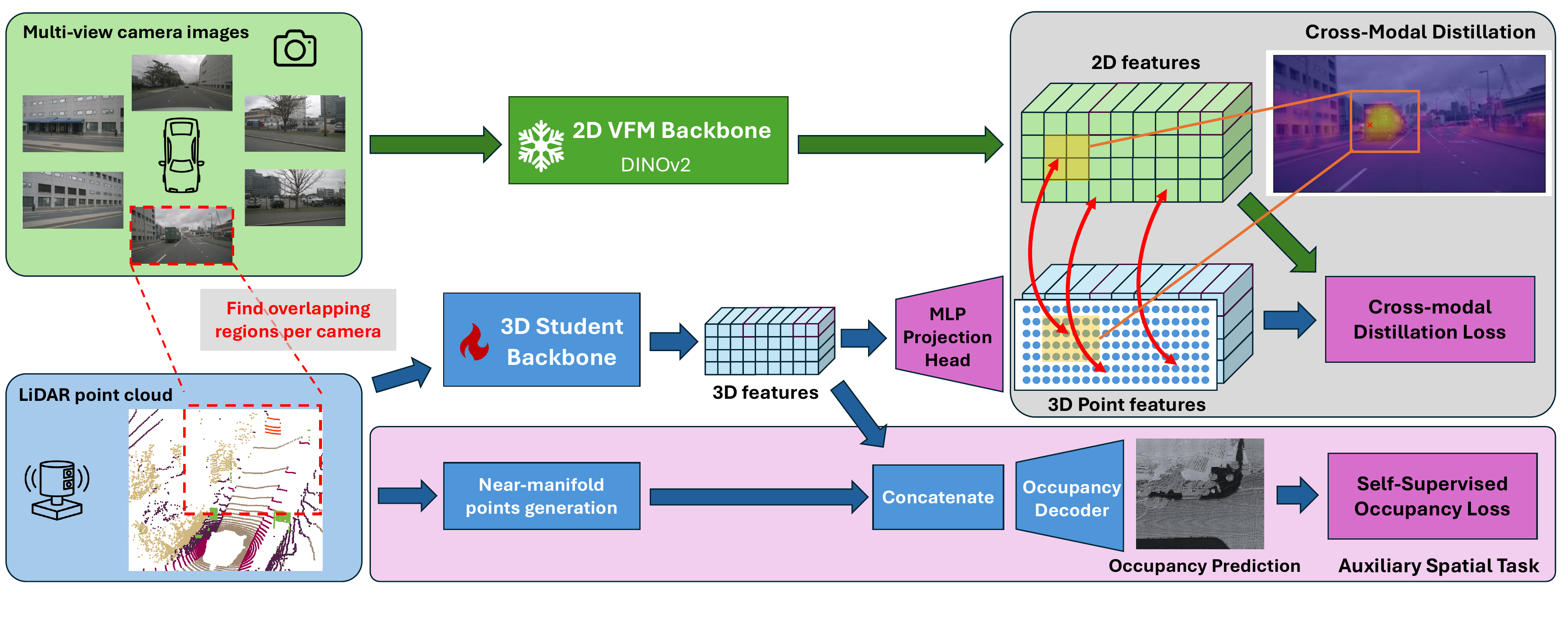}
    \caption{\small Overview of the \textbf{CleverDistiller} framework. Sensor calibration is used to associate 3D points with image regions. Features are extracted using modality-specific backbones. A cross-modal KD loss distills camera features into the 3D backbone via an MLP projection head, while an occupancy loss enforces spatial consistency. The pre-trained 3D backbone is then used for downstream tasks.}
    
    \label{fig:flowchart}
\end{figure*}
\subsection{Cross-modal Learning Objectives}
\label{sec:crossmodal}
Let the student model, $S_{\theta_p} : \mathbb{R}^{N \times 4} \rightarrow \mathbb{R}^{N \times D_p}$ be a 3D backbone with trainable parameters $\theta_p$, which takes LiDAR points as input and outputs $D_p$-dimensional point features. Let $T_{\phi} : \mathbb{R}^{H \times W \times 3} \rightarrow \mathbb{R}^{\frac{H}{S} \times \frac{W}{S} \times D_v}$ be an image backbone with \pretrained parameters $\phi$ that takes images as input and outputs $D_v$-dimensional image features.

The primary objective function of self-supervised knowledge distillation is to transfer knowledge from the trained image backbone $T_{\phi}$ to the 3D backbone $S_{\theta_p}$. The point features are obtained as $\boldsymbol{q}_p = \mathcal{H}_p (S_{\theta_p} (\mathbf{p}))$ and the image features as $\boldsymbol{q}_v = T_{\phi}(\mathbf{I})$, where $\mathcal{H}_p: \mathbb{R}^{D_p} \rightarrow \mathbb{R}^{D_v}$ is a projection head that transforms the point features to the same dimensions as the VFM teacher.
Both features, $q_p$ and $q_v$ are $\ell_2$-normalized. Inspired by \scalr~\cite{scalr}, we use a simple pairwise cosine similarity loss given by,
\begin{equation}
\mathcal{L}_{\mathrm{distillation}}(\boldsymbol{q}_p, \boldsymbol{q}_v) = \frac{1}{N} \sum_{i \in { 1, \ldots, N}} \Vert \boldsymbol{q}_{p,i} - \boldsymbol{q}_{v,\rho(i)} \Vert_2.
\end{equation}

\noindent Such a loss is preferable to a contrastive loss as it avoids the complexities of careful negative sample selection, training sensitivity to batch composition and self-conflict issues while maintaining robust feature alignment. In a multi-camera setup, the above loss is applied to each camera view using only those points that are visible in that camera view.

\subsection{Improving the Projection Head}
\label{sec:fixing_head}

\begin{figure}[t]
  \begin{minipage}[t]{.48\linewidth}
    \centering
    \includegraphics[trim=10 10 0 10, clip, width=\linewidth]{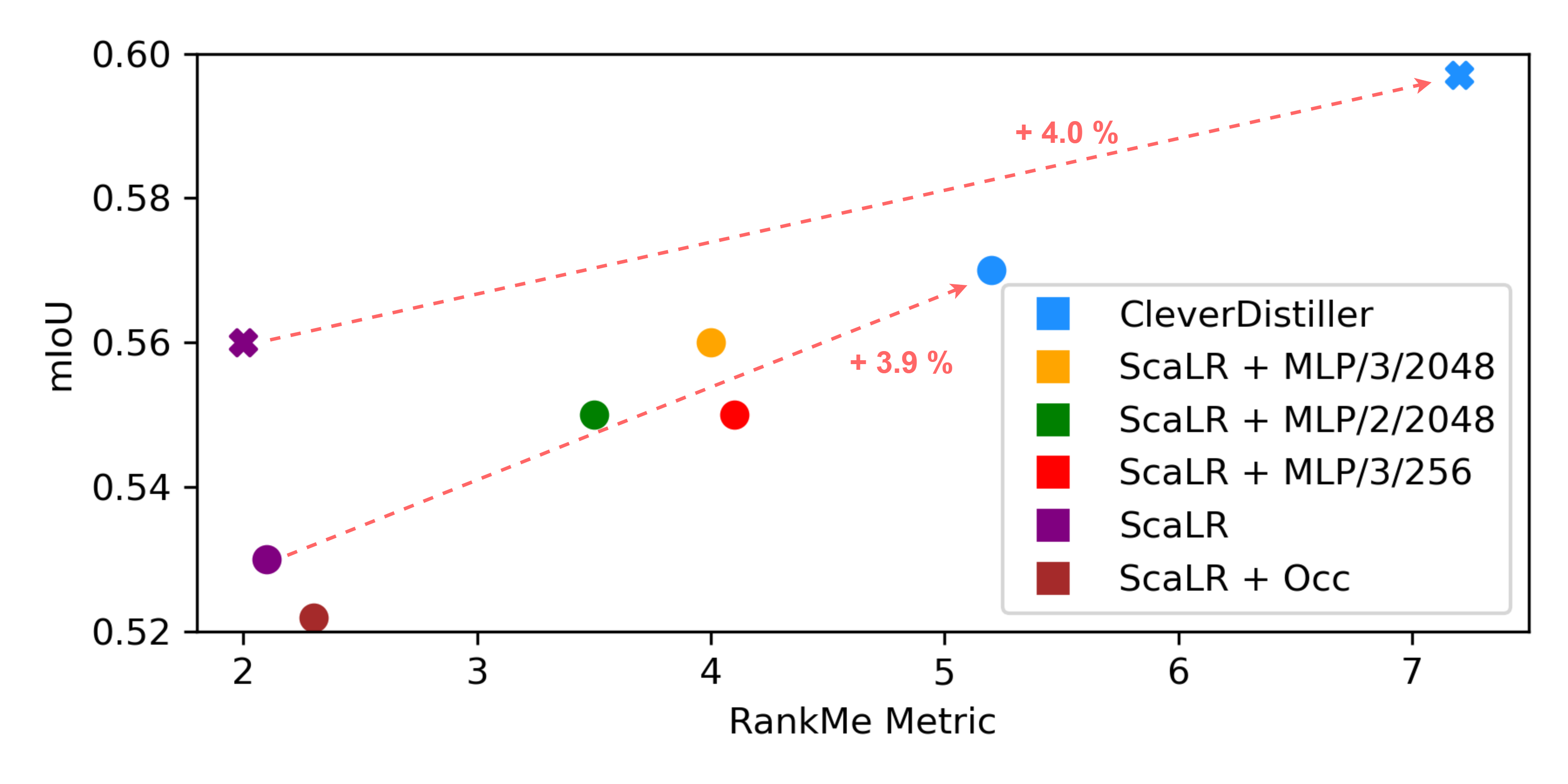}
    \captionof{figure}{\small MinkUNet performance (1\% finetuning) on nuScenes dataset vs RankMe metric of the 3D backbone features. The colors refer to the distillation method, the image teachers are denoted as $\bullet$ for ViT-S/14 and \textbf{$\times$} for ViT-B/14.}
    \label{fig:rank_vs_perf}
  \end{minipage}\hfill
  \begin{minipage}[t]{.48\linewidth}
    \vspace*{-2.8cm}
    \setlength{\tabcolsep}{2.5pt}
    \centering
    \scriptsize
    \begin{tabular}{llccccc}
    \toprule
    \multirow{2}{*}{\makecell{MLP\\Layers.}} & \multirow{2}{*}{\makecell{Hidden\\dim.}} & \multirow{2}{*}{RankMe} & \multicolumn{2}{c}{nuScenes} & KITTI & Waymo \\
    \cmidrule(lr){4-5} \cmidrule(lr){6-6} \cmidrule(lr){7-7}
     & & &  LP & 1\% & 1\% & 1\% \\
    \midrule
    1 (Linear) & - & 1.89 & 40.76 & 53.01 & 44.42 & 47.98 \\
    2 & 2048 & 2.94 & 45.46 & 54.38 & 50.09 & 48.19 \\
    3 & 256 & 4.18 & 44.79 & 54.80 & 50.10 & 48.93 \\
    \cellcolor{white!60!lightgray}3 & \cellcolor{white!60!lightgray}2048 & \cellcolor{white!60!lightgray}\textbf{4.24} & \cellcolor{white!60!lightgray}\textbf{46.36} & \cellcolor{white!60!lightgray}\textbf{55.01} & \cellcolor{white!60!lightgray}\textbf{50.15} & \cellcolor{white!60!lightgray}\textbf{49.08} \\
    \bottomrule
    \end{tabular}
    \vspace{0.5cm}
    \captionof{table}{\small Comparison of different MLP architectures with the PT-v3~\cite{ptv3} 3D backbone model. The MLP projection improves performance as well as the RankMe metric. All scores in mIoU.}
    \label{table:ablation_mlp}
  \end{minipage}
\end{figure}

We investigate the role of the projection head by evaluating the informativeness of the 3D backbone features using the RankMe metric~\cite{rankme}. RankMe is computed \textit{after} self-supervised distillation and \textit{before} any fine-tuning with labels, enabling us to assess the information content of the learned representations in a label-free setting. Originally proposed for evaluating self-supervised models without labels, RankMe serves as a proxy for representational quality and potential downstream performance. We find that this relationship holds in our context as well: higher RankMe scores consistently correlate with better downstream performance after fine-tuning (see \cref{fig:rank_vs_perf,table:ablation_mlp}). This shows that RankMe can help identify design choices in distillation setting, such as importance of projection head or auxiliary tasks (\cref{table:ablation_components}) that yield more informative features for initialization and task-specific training.

Every other method we compare with \cite{scalr,superflow,slidr,xu2025limoe,seal} uses a linear projection head 
 and we find that this results in less informative 3D backbone features (see \cref{table:ablation_mlp}). Using an MLP projection head improves the informativeness of these features as evidenced by the improved RankMe metric. Consequently, this results in significant downstream performance improvements (see \cref{table:ablation_mlp}). A deeper MLP with larger hidden dimensions results in the most informative backbone features and best downstream performance. A complex MLP projection head adds only a minimal increase in computational cost ($\approx9\%$ more peak GPU memory and compute cost). Based on this analysis, we use a 3-layer MLP projection head with a similar formulation as the MLP head used while pre-training the DINOv2 model, which yields the best results (see \cref{table:ablation_mlp,fig:rank_vs_perf}). While the MLP outputs features that align well with the image featureenables the backbone (before the MLP) to retain additional 3D modality-specific information. This is evident in~\cref{table:ablation_components} comparing results with and without occupancy the MLP: with the MLP, the features retain significantly richer information content, as reflected by a higher RankMe metric.

\subsection{Auxiliary Spatial Occupancy Task}
\label{sec:occupancy}

Cross-modal distillation alone does not encourage the model to learn spatially rich information, unique to \lidar that can be useful for downstream tasks. To encourage the learning of spatial features, we incorporate an auxiliary spatial task. Combined with an MLP projection head, this controls the spatial information retained in the backbone features. While many works prioritize segmentation tasks where semantic knowledge is sufficient, tasks such as 3D object detection require a deeper spatial reasoning. We experimented with auxiliary tasks like scene flow estimation and occupancy prediction, ultimately selecting occupancy as the most effective (see supplementary \cref{sec:temporal}).

Occupancy prediction injects spatial and geometric information of the objects into the learned features, which we expect to be useful for tasks such as 3D object detection. The occupancy prediction loss is formulated based on ALSO \cite{also}. We observe that incorporating the occupancy task improves the informativeness of the backbone features by enabling it to learn additional information complementary to those distilled from the VFM features (see improvement on RankMe metric when adding the occupancy task in \cref{table:ablation_components}). This further boosts downstream performance in various tasks. The semantic information from the 2D backbone combined with the spatial knowledge encouraged by occupancy prediction task enable the model to understand and interpret complex 3D environments better. This also makes the model more robust to domain shifts and \lidar corruption (see \cref{tab:robust}).

Following ALSO \cite{also}, for each 3D point \( p \) sampled on the surface, we create two query points: \( q_{\text{front}} = p - \delta p \) and \( q_{\text{behind}} = p + \delta p \), where \( \delta p \) is a small distance. We consider \( q_{\text{front}} \) to be empty and \( q_{\text{behind}} \) to be filled. Additionally, a third empty query point \( q_{\text{sight}} \) is randomly picked in the segment between the sensor and \( p \). Importantly, the occupancy prediction is \textit{not} performed on the original surface points themselves but on these generated query points An occupancy decoder head takes a concatenation of the 3D backbone features of the surface point and the 3D coordinates of the query points to predict the occupancy and intensity of the query points. The occupancy loss consists of a binary cross-entropy loss $\mathcal{L}_{\mathrm{O}}$ based on the occupancy classification and a reconstruction MSE loss $\mathcal{L}_{\mathrm{Int}}$ based on the intensity prediction and we use $\lambda=1$ similar to ALSO as $\mathcal{L}_{\mathrm{occ}} = \mathcal{L}_{\mathrm{O}} + \lambda \mathcal{L}_{\mathrm{Int}}$. 

Overall, we combine two simple but significant design choices. First, we use an MLP projection head and cosine similarity loss. Second, we add a complementary auxiliary spatial task. Combined, these improve the informativeness and spatial consistency of the 3D backbone features as well as downstream performance. This combination of design choices leads to a more robust and efficient cross-modal distillation framework.


\section{Results}
\subsection{Experiments setup}
\label{sec: expt_setup}

\paragraph{Pre-training:} \emph{CleverDistiller} is implemented using the Pointcept framework~\cite{pointcept2023}. Consistent with prior works we employ MinkUNet-34~\cite{choy20194minkunet} as the 3D backbone and DINOv2~\cite{dinov2} (with ViT backbones~\cite{vit}) as the 2D backbone, distilling from three variants: small (S), base (B) and large (L). Additionally, we experiment with SOTA 3D backbone, PointTransformer-V3 (PT-v3) \cite{ptv3}, to investigate if our method scales to larger and newer architectures, showing that CleverDistiller can work with any point-based 3D backbone. For PT-v3, we use the default configuration and only increase the decoder dimensions to 256. The framework is \pretrained on nuScenes using 4 NVIDIA A100-80GB GPUs for 50 and 100 epochs. We follow the same procedure to pre-train both \scalr and CleverDistiller, while the results for other methods are based on results reported in their works~\cite{ppkt,slidr,seal,superflow,xu2025limoe}. Further training setup is provided in \cref{sec:hyperparam}.  
Though most cross-modal image-to-\lidar distillation works claim to pre-train for only 50 epochs, some recent works such as SuperFlow \cite{superflow} utilize 3 \lidar sweeps per sample, increasing the effective length and compute cost of the training by a factor of three (effectively 150 epochs). We characterize this using effective epochs in \cref{tab:minkunet_main_table}. For fairer comparison, we show results of our method for both 50 and 100 epochs of pre-training. CleverDistiller$^{50}$ provides a direct comparison to previous methods that used only one \lidar sweep per sample while CleverDistiller$^{100}$ serves as a more efficient comparison to SuperFlow, ScaLR, and LiMoE+SuperFlow. We experiment on a total of nine datasets in our experiments, namely: nuScenes~\cite{caesar2020nuscenes}, SemanticKITTI~\cite{behley2019semantickitti} (SKITTI), KITTI~\cite{kitti}, Waymo~\cite{sun2020waymo}, ScribbleKITTI~\cite{unal2022scribble}, RELLIS-3D~\cite{jiang2021rellis}, SemanticSTF~\cite{xiao20233d}, and DAPS-3D~\cite{klokov2023daps3d} and nuScenes-C from Robo3D~\cite{kong2023robo3d}. 

\paragraph{Fine tuning:} Pre-trained \lidar backbone is evaluated on a range of downstream tasks on various datasets. Firstly, we conduct segmentation experiments on nuScenes using linear probing (LP) and limited data finetuning scenarios using 4 GPUs for 100 epochs. Further fine-tuning details are provided in \cref{sec:evaluation_setup}. For domain generalization study using Robo3D, we follow the same protocol as previous works. We also test our features on the 3DOD task using the KITTI and Waymo datasets. For 3DOD, we use weights obtained during pre-training on nuScenes. We substitute PointRCNN~\cite{shi2019pointrcnn} backbone with MinkUNet (following SLiDR approach). Finally, we train for 80 epochs on partial data for Waymo and KITTI. 

\paragraph{Evaluation Protocols:} Following standard conventions, we report the Intersection-over-Union (IoU) on each semantic class and mean IoU (mIoU) over all classes for downstream tasks. For 3D robustness evaluations, we follow Robo3D \cite{kong2023robo3d} and report the mean Corruption Error (mCE) and mean Resilience Rate (mRR).

\subsection{Ablation studies}
\label{sec:ablation_studies}

We conduct ablation studies for the two proposed components, the MLP projection head and the auxiliary occupancy task. We pre-train a MinkUNet-34 and PT-V3 model for 100 epochs using the ViT-S/14 variant of DINOv2. We use the 3-layer MLP head selected in \cref{sec:fixing_head} and experimented with different weights for the occupancy loss. From \cref{table:ablation_occ_weight}, we see that a weight of $0.05$ is sufficient to incorporate additional spatial information. Then, we ablate the best versions of the two components \cref{table:ablation_components}. We observe that the addition of the MLP (c) improves the informativeness of the backbone features (RankMe metric) and provides a significant performance boost compared to the baseline (a) in \cref{table:ablation_components}. The occupancy task (d) provides an orthogonal improvement that further improves the features informativeness. Note that the MLP projection head enhances the effectiveness of the occupancy task, as it enables the backbone features to learn additional \lidar-specific spatial information which may not align with the purely semantic image features.


\begin{table}[t!]
\begin{minipage}[t]{0.43\textwidth}
\scriptsize
\setlength{\tabcolsep}{5pt}
\centering
\begin{tabular}{lcccccc}
\toprule
\multirow{2}{*}{\makecell{Occupancy\\loss weight}} & \multicolumn{2}{c}{nuScenes} & SKITTI & Waymo \\
\cmidrule(lr){2-3} \cmidrule(lr){4-4} \cmidrule(lr){5-5}
 & LP & 1\% & 1\% & 1\% \\
\midrule
0.0 & 55.72 & 61.79 & 56.30 & 60.78 \\
0.01 & 56.49 & 61.40 & 57.01 & 62.14 \\
\cellcolor{white!60!lightgray}0.05 & \cellcolor{white!60!lightgray}\textbf{58.49} & \cellcolor{white!60!lightgray}\textbf{62.54} & \cellcolor{white!60!lightgray}\textbf{62.19} & \cellcolor{white!60!lightgray}61.95 \\
0.2 & 57.39 & 61.65 & 58.96 & \textbf{62.39} \\
1.0 & 53.89 & 58.99 & 56.78 & 62.30 \\
\bottomrule
\end{tabular}
\vspace{0.2cm}
\caption{\small Comparison of different occupancy loss weights with PTV3~\cite{ptv3}, results in mIoU.}
\label{table:ablation_occ_weight}
\end{minipage}
\hspace{0.02\textwidth}
\begin{minipage}[t]{0.53\textwidth}
\scriptsize
\centering
\setlength{\tabcolsep}{1.5pt}
\begin{tabular}{lcccccccc}
\toprule
\multirow{2}{*}{\#} & \multirow{2}{*}{MLP} & \multirow{2}{*}{Occ} & \multirow{2}{*}{\makecell{Matrix\\Rank}} & \multirow{2}{*}{RankMe} & \multicolumn{2}{c}{nuScenes} & SKITTI & Waymo \\
\cmidrule(lr){6-7} \cmidrule(lr){8-8} \cmidrule(lr){9-9}
& & & & & LP & 1\% & 1\% & 1\% \\
\midrule
(a) & \textcolor{red}{\xmark} & \textcolor{red}{\xmark} & 22 & 1.89 & 40.76 & 53.01 & 44.42& 47.98 \\
(b) & \textcolor{red}{\xmark} & \textcolor{green}{\cmark} & 25 & 2.33 & 41.67 & 52.16 & 48.19 & 48.11 \\
(c) & \textcolor{green}{\cmark} & \textcolor{red}{\xmark} & 33 & 4.24 & 46.36 & 55.01 & 50.15 & 49.08 \\
\cellcolor{white!60!lightgray}(d) & \cellcolor{white!60!lightgray}\textcolor{green}{\cmark} & \cellcolor{white!60!lightgray}\textcolor{green}{\cmark} & \cellcolor{white!60!lightgray}41 & \cellcolor{white!60!lightgray}5.27 & \cellcolor{white!60!lightgray}49.81 & \cellcolor{white!60!lightgray}56.90 & \cellcolor{white!60!lightgray}50.59 & \cellcolor{white!60!lightgray}50.99 \\
\bottomrule
\end{tabular}
\vspace{0.33cm}
\caption{\small Ablation study of different components with MinkUNet-34~\cite{choy20194minkunet}, results in mIoU.}
\label{table:ablation_components}
\end{minipage}
\end{table}

\subsection{Performance Comparison}
\label{sec:mainresults}
%
\begin{table*}[t!]
\centering
\scriptsize

\renewcommand{\arraystretch}{0.7} 
\setlength{\tabcolsep}{4.5pt}
\resizebox{\linewidth}{!}{
\begin{tabular}{lcccccccccccc}

\toprule
\multirow{2}{*}{{\textbf{Method}}} &  \multirow{2}{*}{{\textbf{Venue}}} &  \multirow{2}{*}{{\textbf{Distill}}} &  \multirow{2}{*}{{\textbf{\begin{tabular}[c]{@{}c@{}}S.P.\end{tabular}}}} &   \multirow{2}{*}{{\textbf{\begin{tabular}[c]{@{}c@{}}Effective\\ Epochs\end{tabular}}}} & \multicolumn{6}{c}{\textbf{nuScenes}} &  {\textbf{SKITTI}} & \multicolumn{1}{c}{\textbf{Waymo}} \\

\cmidrule(lr){6-11} \cmidrule(lr){12-12} \cmidrule(lr){13-13} 

 {} &  {} &  {} &  {} &  {} & \textbf{LP} & \textbf{1\%} & \textbf{5\%} & \textbf{10\%} & \textbf{25\%} &  \textbf{Full}  &  \textbf{1\%} & \textbf{1\%} \\ 

\hline

Random & - & - & {\textcolor{red}{\xmark}} & - & {8.10} & {30.30} & {47.84} & {56.15} & {65.48} & {74.66} & {39.50} & 39.41 \\ 

\arrayrulecolor{lightgray} \hline

 {PPKT~\cite{ppkt}} &  {arXiv’21} &  {{\color[HTML]{2FC061} ViT-S}} &  {\textcolor{red}{\xmark}} &  {50} &  {38.60} &  {40.60} &  {52.06} &  {59.99} &  {65.76} &  {73.97} &  {43.25} & 47.44 \\ 

\hline

 {SLidR~\cite{slidr}} &  {CVPR’22} &  {{\color[HTML]{2FC061} ViT-S}} &  {\textcolor{green}{\cmark}} &  {50} &  {44.70} &  {41.16} &  {53.65} &  {61.47} &  {66.71} &  {74.20} &  {44.67} & 47.57 \\ 

\hline

 {Seal~\cite{seal}} &  {NIPS’23} &  {{\color[HTML]{2FC061} ViT-S}} &  {\textcolor{green}{\cmark}} &  {50} &  {45.16} &  {44.27}   &  {55.13} &  {62.46} &  {67.64} &  {75.58} &  {46.51} & 48.67 \\ 

\hline

 {\textbf{CleverDistiller}$^{50}$} &  {-} &  {{\color[HTML]{2FC061} ViT-S}} &  {\textcolor{red}{\xmark}} &  {50} &  \cellcolor{white!70!SpringGreen}{\textbf{46.02}} &  \cellcolor{white!70!SpringGreen}{\textbf{53.67}} &  \cellcolor{white!70!SpringGreen}{\textbf{62.34}} &  \cellcolor{white!70!SpringGreen}{\textbf{64.04}} &  \cellcolor{white!70!SpringGreen}{\textbf{68.06}} &   \cellcolor{white!70!SpringGreen}{\textbf{77.44}} &  \cellcolor{white!70!SpringGreen}{\textbf{49.58}} & \cellcolor{white!70!SpringGreen}\textbf{50.45} \\ 

\specialrule{.2em}{.1em}{.1em}

 {SuperFlow~\cite{superflow}} &  {ECCV'24} &  {{\color[HTML]{2FC061} ViT-S}} &  {\textcolor{green}{\cmark}} &  {150} &  {46.44} &  {47.81} &  {59.44} &  {64.47} &  {69.20} &  {76.54} &  {47.97} & 49.94 \\ 

\hline

 {ScaLR~\cite{scalr}} &  {CVPR'24} &  {{\color[HTML]{2FC061} ViT-S}} &  {\textcolor{red}{\xmark}} &  {100} &  {40.76} &  {53.01} &  {61.58} &  {63.47} &  {67.56} &  {74.51} &  {44.42} & 47.98 \\ 

\hline

 {LiMoE~\cite{xu2025limoe}} &  {CVPR'25} &  {\color[HTML]{2FC061}ViT-S} &  {\textcolor{green}{\cmark}} &  {150} &  {48.20} &  {49.60} &  {60.54} &  {65.65} &  \cellcolor{white!0!SpringGreen}{\textbf{71.39}} &  {77.27} &  {49.53} & \cellcolor{white!0!SpringGreen}\textbf{51.42} \\ \hline

 {\textbf{CleverDistiller}$^{100}$} &  {-} &  {{\color[HTML]{2FC061} ViT-S}} &  {\textcolor{red}{\xmark}} &  {100} &  \cellcolor{white!0!SpringGreen}{\textbf{49.81}} &  \cellcolor{white!0!SpringGreen}{\textbf{56.90}} &  \cellcolor{white!0!SpringGreen}{\textbf{64.55}} &  \cellcolor{white!0!SpringGreen}{\textbf{65.92}} &  {70.11} &  \cellcolor{white!0!SpringGreen}{\textbf{77.61}} &  \cellcolor{white!0!SpringGreen}{\textbf{50.59}} & 50.99 \\ 

\arrayrulecolor{black} \midrule \midrule \arrayrulecolor{lightgray}

 {PPKT~\cite{ppkt}} &  {arXiv’21} &  {{\color[HTML]{FD6864} ViT-B}} &  {\textcolor{red}{\xmark}} &  {50} &  {39.95} &  {40.91} &  {53.21} &  {60.87} &  {66.22} &  {74.07} &  {44.09} & 47.57 \\ 

\hline

 {SLidR~\cite{slidr}} &  {CVPR’22} &  {{\color[HTML]{FD6864} ViT-B}} &  {\textcolor{green}{\cmark}} &  {50} &  {45.35} &  {41.64} &  {55.83} &  {62.68} &  {67.61} &  {74.98} &  {45.50} & 48.32 \\ 

\hline

 {Seal~\cite{seal}} &  {NIPS’23} &  {{\color[HTML]{FD6864} ViT-B}} &  {\textcolor{green}{\cmark}} &  {50} &  {46.59} &  {45.98}   &  {57.15} &  {62.79} &  {68.18} &  {75.41} &  {47.24} & 48.91 \\ 

\hline

 {\textbf{CleverDistiller}$^{50}$}  &  {-} &  {{\color[HTML]{FD6864} ViT-B}} &  {\textcolor{red}{\xmark}} &  {50} &  \cellcolor{white!90!RubineRed}{\textbf{48.40}} &  \cellcolor{white!90!RubineRed}{\textbf{55.80}} &  \cellcolor{white!90!RubineRed}{\textbf{63.49}} &  \cellcolor{white!90!RubineRed}{\textbf{65.56}} &  \cellcolor{white!90!RubineRed}{\textbf{69.06}} &  \cellcolor{white!90!RubineRed}{\textbf{78.17}} &  \cellcolor{white!90!RubineRed}{\textbf{49.83}} & \cellcolor{white!90!RubineRed}\textbf{52.80} \\ 
\specialrule{.2em}{.1em}{.1em} 

 {SuperFlow~\cite{superflow}} &  {ECCV'24} &  {{\color[HTML]{FD6864} ViT-B}} &  {\textcolor{green}{\cmark}} &  {150} &  {47.66} &  {48.09}   &  {59.66} &  {64.52} &  {69.79} &  {76.57} &  {48.40} & 50.20 \\ 

\hline

 {ScaLR~\cite{scalr}} &  {CVPR'24} &  {{\color[HTML]{FD6864} ViT-B}} &  {\textcolor{red}{\xmark}} &  {100} &  {41.80} &  {55.83} &  {63.46} &  {65.24} &  {68.70} &  {74.76} &  {45.59} & 49.60 \\ 

\hline

 {LiMoE~\cite{xu2025limoe}} &  {CVPR'25}        &  {{\color[HTML]{FD6864} ViT-B}} &  {\textcolor{green}{\cmark}} &  {150} &  {49.07} &  {50.23} &  {61.51} &  {66.17} &  \cellcolor{white!55!RubineRed}{\textbf{71.56}} &  {77.81} &  {50.30} & 51.77 \\ \hline

 {\textbf{CleverDistiller}$^{100}$} &  {-} &  {{\color[HTML]{FD6864} ViT-B}} &  {\textcolor{red}{\xmark}} &  {100} &  \cellcolor{white!55!RubineRed}{\textbf{51.89}} &  \cellcolor{white!55!RubineRed}{\textbf{59.80}} &  \cellcolor{white!55!RubineRed}{\textbf{66.44}} &  \cellcolor{white!55!RubineRed}{\textbf{67.65}} &  {69.53} &  \cellcolor{white!55!RubineRed}{\textbf{78.49}} &  \cellcolor{white!55!RubineRed}{\textbf{51.48}} & \cellcolor{white!55!RubineRed}\textbf{53.56} \\ 

\arrayrulecolor{black} \midrule \midrule \arrayrulecolor{lightgray}

 {PPKT~\cite{ppkt}} &  {arXiv’21} &  {{\color[HTML]{6434FC} ViT-L}} &  {\textcolor{red}{\xmark}} &  {50} &  {41.57} &  {42.05} &  {55.75} &  {61.26} &  {66.88} &  {74.33} &  {45.87} & 47.82 \\ 

\hline

 {SLidR~\cite{slidr}} &  {CVPR’22} &  {{\color[HTML]{6434FC} ViT-L}} &  {\textcolor{green}{\cmark}} &  {50} &  {45.70} &  {42.77} &  {57.45} &  {63.20} &  {68.13} &  {75.51} &  {47.01} & 48.60 \\ 

\hline

 {Seal~\cite{seal}} &  {NIPS’23} &  {{\color[HTML]{6434FC} ViT-L}} &  {\textcolor{green}{\cmark}} &  {50} &  {46.81} &  {46.27} &  {58.14} &  {63.27} &  {68.67} &  {75.66} &  {47.55} & 50.02 \\ 

\hline

 {\textbf{CleverDistiller}$^{50}$} &  {-} &  {{\color[HTML]{6434FC} ViT-L}} &  {\textcolor{red}{\xmark}} &  {50} &   \cellcolor{white!82!Periwinkle}{\textbf{48.32}} &   \cellcolor{white!82!Periwinkle}{\textbf{56.70}} &   \cellcolor{white!82!Periwinkle}{\textbf{64.51}} &  \cellcolor{white!82!Periwinkle}{\textbf{66.25}} &  \cellcolor{white!82!Periwinkle}{\textbf{69.33}} &  \cellcolor{white!82!Periwinkle}{\textbf{78.68}} &  \cellcolor{white!82!Periwinkle}{\textbf{49.86}} & \cellcolor{white!82!Periwinkle}\textbf{53.32} \\ 

\specialrule{.2em}{.1em}{.1em} 

 {SuperFlow~\cite{superflow}} &  {ECCV'24} &  {{\color[HTML]{6434FC} ViT-L}} &  {\textcolor{green}{\cmark}} &  {150} &  {48.01} &  {49.95}   &  {60.72} &  {65.09} &  {70.01} &  {77.19} &  {49.07} & 50.67 \\ 

\hline

 {ScaLR~\cite{scalr}} &  {CVPR'24} &  {{\color[HTML]{6434FC} ViT-L}} &  {\textcolor{red}{\xmark}} &  {100} &  {40.12} &  {55.78} &  {63.28} &  {64.76} &  {68.19} &  {75.09} &  {44.85} & 50.34 \\ 

\hline

 {LiMoE~\cite{xu2025limoe}} &  {CVPR'25} &  {{\color[HTML]{6434FC} ViT-L}}  &  {\textcolor{green}{\cmark}} &  {150} &  {49.35} &  {51.41} &  {62.07} &  {66.64} &  \cellcolor{white!50!Periwinkle}{\textbf{71.59}} &  {77.85} &  {50.69} & 51.93 \\ 

\hline

\textbf{CleverDistiller}$^{100}$ & - & {\color[HTML]{6434FC} ViT-L} & \textcolor{red}{\xmark} & 100 & \cellcolor{white!50!Periwinkle}\textbf{52.45} & \cellcolor{white!50!Periwinkle}\textbf{60.64} & \cellcolor{white!50!Periwinkle}\textbf{67.03} & \cellcolor{white!50!Periwinkle}\textbf{67.29} & 70.45 & \cellcolor{white!50!Periwinkle}\textbf{78.29} & \cellcolor{white!50!Periwinkle}\textbf{52.28} & \cellcolor{white!50!Periwinkle}\textbf{54.83} \\ 

\arrayrulecolor{black} \bottomrule

\end{tabular}
}
\vspace{0.2cm}
\caption{\small Comparison of SOTA pretraining methods pretrained in self-supervised manner on nuScenes and fine-tuned on nuScenes, SemanticKITTI, and Waymo. All methods use MinkUNet-34~\cite{choy20194minkunet} as the backbone for the 3D semantic segmentation task. We also show which 2D VFM backbone was used as a teacher and if the approach uses pseudo semantic priors from off-the shelf segmentation models. LP denotes linear probing with a frozen backbone. All scores in mIoU. For LiMoE~\cite{xu2025limoe}, we use LiMoE+SuperFlow. "S.P." refers to whether the method uses Semantic Priors.}
\label{tab:minkunet_main_table}
\end{table*}

\begin{table}[ht]
\begin{minipage}{0.45\linewidth}
\notsotiny
\setlength{\tabcolsep}{1.2pt}
\centering
\begin{tabular}{lccccccccc}
\toprule
\textbf{Method} & \multicolumn{5}{c}{\textbf{nuScenes}} & \textbf{SKITTI} & \textbf{Waymo} \\
\cmidrule(lr){2-6} \cmidrule(lr){7-7} \cmidrule(lr){8-8}
 & \textbf{LP} & \textbf{1\%} & \textbf{5\%} & \textbf{10\%} & \textbf{25\%} & \textbf{1\%} & \textbf{1\%} \\ 
 \hline
{\color[HTML]{2FC061} ViT-S}  &  &  &  & &  & &  &  \\ 
 ScaLR  &  {44.99} & {62.07} & {69.10} & {72.91} & {75.69} & 56.92 & 60.09 \\ 

 \textbf{Ours} &  \cellcolor{white!70!SpringGreen}{\textbf{58.49}} & \cellcolor{white!70!SpringGreen}{\textbf{62.54}} & \cellcolor{white!70!SpringGreen}\textbf{71.09} & \cellcolor{white!70!SpringGreen}{\textbf{73.89}} & \cellcolor{white!70!SpringGreen}\textbf{76.64} & \cellcolor{white!70!SpringGreen}\textbf{61.95} & \cellcolor{white!70!SpringGreen}\textbf{62.19} \\ 
\midrule
{\color[HTML]{FD6864} ViT-B}  &  &  &  & &  & &  &  \\ 
  ScaLR &  {41.62} & {63.54} & {71.74} & {73.84} & {76.56} & 60.74 & 57.17 \\ 
    \textbf{Ours} &  \cellcolor{white!90!RubineRed}\textbf{60.54} & \cellcolor{white!90!RubineRed}\textbf{63.93} & \cellcolor{white!90!RubineRed}\textbf{72.27} & \cellcolor{white!90!RubineRed}\textbf{74.11} & \cellcolor{white!90!RubineRed}\textbf{76.93} &\cellcolor{white!90!RubineRed}\textbf{63.31} & \cellcolor{white!90!RubineRed}\textbf{62.20}
     \\ 
\midrule 
{\color[HTML]{6434FC} ViT-L}  &  &  &  & &  & &  &  \\ 
  ScaLR &  {45.54} & {66.17} & {72.34} & {74.96} & {77.56} & 58.48 & 61.47 \\ 
    \textbf{Ours} &  \cellcolor{white!82!Periwinkle}\textbf{61.92} & \cellcolor{white!82!Periwinkle}\textbf{66.18} & \cellcolor{white!82!Periwinkle}\textbf{73.13} & \cellcolor{white!82!Periwinkle}\textbf{75.18} & \cellcolor{white!82!Periwinkle}\textbf{77.57} & \cellcolor{white!82!Periwinkle}\textbf{62.23} & \cellcolor{white!82!Periwinkle}\textbf{65.94} \\
\bottomrule
\end{tabular}
\vspace{0.2cm}
\caption{Comparisons of SOTA methods pretrained on nuScenes and fine-tuned on datasets subset, using PTV3~\cite{ptv3} for 3D semantic segmentation. All scores in mIoU.}
\label{tab:ptv3_main_table}
\end{minipage}
\hfill
\begin{minipage}{0.5\linewidth}
\centering
\scriptsize
\setlength{\tabcolsep}{1.2pt}
\notsotiny
\begin{tabular}{lcccccccc}
\toprule
\multirow{2}{*}{\textbf{Method}} & \multicolumn{2}{c}{\textbf{ScriKITTI}} & \multicolumn{2}{c}{\textbf{Rellis-3D}} & \multicolumn{2}{c}{\textbf{SemSTF}} & \multicolumn{2}{c}{\textbf{DAPS-3D}} \\
\cmidrule(lr){2-3} \cmidrule(lr){4-5} \cmidrule(lr){6-7} \cmidrule(lr){8-9}
& \multicolumn{1}{l}{\textbf{1\%}} & {\textbf{10\%}} & \multicolumn{1}{l}{\textbf{1\%}} & {\textbf{10\%}} & \multicolumn{1}{l}{\textbf{50\%}} & {\textbf{100\%}} & \multicolumn{1}{l}{\textbf{50\%}} & \textbf{100\%} \\ 
\midrule
Random & {23.81} & 47.60 & {38.46} & 53.60 & {48.03} & 48.15 & 74.32 & 79.38 \\ 
\arrayrulecolor{lightgray} \hline
PPKT$^{50}$ & {36.50} & 51.67 & {49.71} & 54.33 & {50.92} & 54.69 & 78.90 & 84.00 \\ 
\hline
SLidR$^{50}$ & {39.60} & 50.45 & {49.75} & 54.57 & {52.01} & 54.35 & 81.00 & 85.40 \\ 
\hline
Seal$^{50}$ & {40.64} & 52.77 & {51.09} & 55.03 & {53.46} & \cellcolor{white!70!SpringGreen}\textbf{55.36} & {81.88} & {85.90} \\ 
\hline
\textbf{Ours}$^{50}$ & \cellcolor{white!70!SpringGreen}{\textbf{40.77}} & \cellcolor{white!70!SpringGreen}\textbf{53.31} & \cellcolor{white!70!SpringGreen}{\textbf{56.23}} & \cellcolor{white!70!SpringGreen}\textbf{56.60} & \cellcolor{white!70!SpringGreen}{\textbf{53.77}} & 54.72 & \cellcolor{white!70!SpringGreen}\textbf{82.58} & \cellcolor{white!70!SpringGreen}\textbf{86.15} \\ 
\arrayrulecolor{black} \midrule \midrule \arrayrulecolor{lightgray}
ScaLR$^{100}$ & {36.45} & 49.16 & { 47.91  } & 48.86 & {52.10 } & 54.40 & - & - \\ 
\hline
SF$^{150}$ & {42.70} & 54.00 & {52.83} & 55.71 & {54.72} & 56.57 & 82.43 & 86.21 \\ 
\hline
LiMoE$^{150}$ & {43.95} & 55.96 & {53.74} & 56.67 & \cellcolor{white!20!SpringGreen}{\textbf{55.60}} & \cellcolor{white!20!SpringGreen}\textbf{57.31  } & 83.24 & 86.68 \\ \hline
\textbf{Ours}$^{100}$ & \cellcolor{white!20!SpringGreen}{\textbf{44.03}} & \cellcolor{white!20!SpringGreen}\textbf{56.70} & \cellcolor{white!20!SpringGreen}{\textbf{58.35}} & \cellcolor{white!20!SpringGreen}\textbf{60.92} & {53.99} & 55.66 & \cellcolor{white!20!SpringGreen}\textbf{83.06} & \cellcolor{white!20!SpringGreen}\textbf{87.95} \\
\arrayrulecolor{black} \bottomrule
\end{tabular}
\vspace{0.2cm}
\caption{Domain generalization benchmark. Models are pre-trained on nuScenes and finetuned on different datasets. (all scores in mIoU)}
\label{tab:generalization}
\end{minipage}
\end{table}

As shown in \cref{tab:minkunet_main_table} and \cref{tab:ptv3_main_table}, CleverDistiller achieves state-of-the-art performance across all fine-tuning tasks, as well as in linear probing, on both MinkUnet~\cite{choy20194minkunet} and PTV3~\cite{ptv3}. In particular, when fine-tuning with limited data on nuScenes, we observe substantial improvements of up to 10\% mIoU in comparison to baselines. CleverDistiller also consistently outperforms competing methods in domain transfer scenarios, such as SKITTI and Waymo. To the best of our knowledge, we include all recently published and accepted methods (including \cite{xu2025limoe}), and demonstrate that high performance can be achieved through simple design choices, without relying on semantic priors or contrastive losses.

The performance of CleverDistiller highlights strong generalization of the learned representations, making it well-suited for settings with limited annotated data. The results also confirm that models pretrained via representation learning significantly outperform those initialized randomly, underscoring the importance of data-driven pretraining. Furthermore, distillation from larger 2D networks consistently improves downstream performance, aligning with findings from prior work. We further provide additional qualitative results, feature visualizations, and class-specific analyses in the supplementary (\cref{sec: app_classwise_results,sec:visualization}).





\subsection{Domain generalization}
\label{sec:domaingeneral}

In domain generalization experiments, we evaluate all methods pre-trained on nuScenes across four diverse datasets as shown in~\cref{tab:generalization}. We fine-tune the models on different subsets of data from the new dataset using the fine-tuning protocol described in \cref{sec: expt_setup}. Each dataset poses distinct challenges, with some originating outside the autonomous driving domain. Notably, ScriKITTI features extremely sparse scribble-level annotations, yet our method still achieves the best performance among all approaches, highlighting its ability to learn effectively from limited supervision. Rellis-3D, in contrast, is an off-road mobile robotics dataset with a significant domain shift from nuScenes in terms of terrain, sensor configuration, and scene layout. Despite this, our method adapts well and continues to outperform others, indicating strong generalization across both structured and unstructured environments. The results on SemSTF and DAPS-3D further support the robustness of our approach in diverse settings.

\subsection{Robustness}
\label{sec:robustness}
We compare our method against SOTA approaches on the Robo3D benchmark~\cite{kong2023robo3d} presented in \cref{tab:robust} based on the linear probing models. CleverDistiller achieves superior performance, particularly in corrupted settings where other methods degrade significantly for example, when LiDAR data is sparse. It consistently shows improved robustness over random initialization and competing methods, obtaining the lowest mean Corruption Error (mCE) in linear probing (robustness results based on full data finetuned models are provided in the supplementary, \cref{sec:robust_full}). Our method also achieves the highest mean Relative Robustness (mRR) across all trials, indicating strong resilience to various corruptions. In the Fog scenario, for instance, CleverDistiller reaches an mIoU of 43.96, outperforming the previous best of 40.35. The most pronounced improvement is observed in the cross-sensor corruption scenario, characterized by extremely sparse LiDAR input. Here, CleverDistiller performs up to 20\% points better in linear probing accuracy compared to other methods.

Although LiMoE~\cite{xu2025limoe} (very recent unpublished work) slightly outperforms our method in some cases, CleverDistiller remains more robust on average, achieving lower mCE and higher mRR. Notably, our method is simpler and requires significantly lesser training compute than the mixture-of-experts model, LiMoE, highlighting its efficiency and effectiveness.

\begin{table*}[t!]
\scriptsize
\renewcommand{\arraystretch}{0.5} 
\setlength{\tabcolsep}{2.5pt}
\centering
\begin{tabular}{cccccccccccccc}
\toprule
 \multirow{2}{*}{Method} & \multirow{2}{*}{Backbone} & {\multirow{2}{*}{mCE $\downarrow$}} & {\multirow{2}{*}{mRR $\uparrow$}} & \multicolumn{9}{c}{mIoU $\uparrow$} \\
\cmidrule(lr){5-13}
 \multicolumn{2}{l}{} & {} & {} & {Fog}   & {Rain} & {Snow}  & {Blur} & {Beam} & {Cross} & {Echo} & {Sensor} & Avg Full \\ 
 \midrule


{PPKT$^{50}$~\cite{ppkt}} & MinkU-34 & 183.44 & 78.15 & {30.65} & {35.42} & {28.12} & {29.21} & {32.82} & {19.52} & {28.01} & {20.71} & 28.06 \\

\cmidrule(lr){1-13}

{SLidR$^{50}$~\cite{slidr}} & MinkU-34 & 179.38 & 77.18 & {34.88} & {38.09} & {32.64} & {26.44} & {33.73} & {20.81} & {31.54} & {21.44} & 29.95 \\

\cmidrule(lr){1-13}

{Seal$^{50}$~\cite{seal}} & MinkU-34 & 166.18 & 75.38 & {37.33} & {42.77} & {29.93} & {37.73} & {40.32} & {20.31} & {37.73} & {24.94} & 33.88 \\

\cmidrule(lr){1-13}

{\textbf{CleverDistiller}$^{50}$} & MinkU-34 & \cellcolor{white!70!SpringGreen}\textbf{157.50}& \cellcolor{white!70!SpringGreen}\textbf{84.84}& \cellcolor{white!70!SpringGreen}{\textbf{42.45}} & \cellcolor{white!70!SpringGreen}{\textbf{43.45}} & \cellcolor{white!70!SpringGreen}{\textbf{38.77}} & \cellcolor{white!70!SpringGreen}{\textbf{37.85}} & \cellcolor{white!70!SpringGreen}{\textbf{41.01}} & \cellcolor{white!70!SpringGreen}{\textbf{43.11}} & \cellcolor{white!70!SpringGreen}{\textbf{39.17}} & \cellcolor{white!70!SpringGreen}{\textbf{26.55}} & \cellcolor{white!70!SpringGreen}\textbf{39.05}\\ 

\cmidrule[2pt](lr){1-13}

{SuperFlow$^{150}$~\cite{superflow}} & MinkU-34 & 161.78 & 75.52 & {37.59} & {43.42} & {37.60} & {39.57} & {41.40} & {23.64} & {38.03} & {26.69} & 35.99 \\

\cmidrule(lr){1-13}

{Scalr$^{100}$~\cite{scalr}} & MinkU-34 & 173.18 & 78.91 & {37.55} & {37.96} & {36.29} & {33.64} & {33.06} & {23.01} & {33.62} & {23.70} & 33.59 \\

\cmidrule(lr){1-13}
                    
{LiMoE$^{150}$} & MinkU-34 & 155.77 & 78.23 & {40.35} & {45.28} & {39.14} & \cellcolor{white!0!SpringGreen}{\textbf{42.10}} & \cellcolor{white!0!SpringGreen}{\textbf{44.21}} & {27.33} & {39.20} & \cellcolor{white!0!SpringGreen}{\textbf{29.49}} & 38.39 \\
\cmidrule(lr){1-13}

{\textbf{CleverDistiller}$^{100}$} & MinkU-34 & \cellcolor{white!0!SpringGreen}\textbf{151.21} & \cellcolor{white!0!SpringGreen}\textbf{89.99} & \cellcolor{white!0!SpringGreen}{\textbf{43.96}} & \cellcolor{white!0!SpringGreen}{\textbf{46.91}} & \cellcolor{white!0!SpringGreen}{\textbf{41.20}} & {41.05} & {42.15} & \cellcolor{white!0!SpringGreen}\cellcolor{white!0!SpringGreen}{\textbf{45.67}} & \cellcolor{white!0!SpringGreen}{\textbf{41.30}} & {28.85} & \cellcolor{white!0!SpringGreen}\textbf{41.39 }\\

\arrayrulecolor{black} \bottomrule
\end{tabular}
\vspace{0.2cm}
\caption{\small 3D robustness study under corruption and sensor failure scenarios in the nuScenes-C dataset from the Robo3D benchmark~\cite{kong2023robo3d}. This evaluation is based on the linear probing model trained on frozen MinkUNet-34 backbones obtained using different knowledge distillation methods. All mCE (↓), mRR (↑), and mIoU (↑) scores are given in percentage (\%).}
\label{tab:robust}
\end{table*}





\section{Conclusions}
\label{sec:conclusions}
We introduce CleverDistiller, a simple yet effective framework for cross-modal distillation from 2D vision foundation models (VFMs) to 3D LiDAR models. Unlike prior work relying on complex losses or pseudo-labels, we use a direct feature similarity loss and an MLP projection head to improve semantic information transfer. We further enhance spatial reasoning by incorporating an auxiliary task of occupancy prediction. CleverDistiller achieves SOTA performance with up to 10\% mIoU gains in low-data regimes.

\section*{Acknowledgments}

This work was partially supported by the Wallenberg AI, Autonomous Systems and Software Program (WASP) funded by the Knut and Alice Wallenberg Foundation.


\bibliographystyle{plain}
\bibliography{references}
\clearpage

\section*{Supplementary Material}
\setcounter{table}{0}
\setcounter{page}{1}
\renewcommand{\thetable}{A\arabic{table}}
\setcounter{figure}{0}
\renewcommand{\thefigure}{A\arabic{figure}}
\renewcommand{\thesection}{A\arabic{section}}
\setcounter{section}{0}
\section{Limitations}
Although our method is robust to data corruptions, we use well calibrated data for the distillation and use the calibration information to map points to pixels in the image. This results in the limitation that our method is sensitive to sensor decalibration: small misalignment is tolerable due to ViT’s patch-based structure, but large ones can cause incorrect matching of points to pixels and degrade performance. This is partially addressed by using pointcloud augmentations during distillation (such as, jittering - randomly moving points in the point cloud by a small amount) but it is still not enough if large decalibration occurs.

We demonstrate extensive experiments using the DINOv2 teacher and using the two \lidar backbones, MinkUNet-34 \cite{choy20194minkunet} and PointTransformer-V3 \cite{ptv3}. MinkUNet-34 is used in benchmark experiments in prior works on cross-modal knowledge distillation. We are the first to investigate cross-modal knowledge distillation for the PointTransformer-V3 backbone, a recent state-of-the-art \lidar backbone. However, one limitation is that we do not study different teacher vision models. Different vision models produce features containing different information - studying the properties of the \lidar backbone distilled from different teacher vision models or developing methods to distill from multiple teachers are interesting future directions.

\section{Hyperparameters setup}
\label{sec:hyperparam}

We implement both \scalr and our method, CleverDistiller in the Pointcept framework \cite{pointcept2023}. We use the standard training configuration for PointTransformerV3 and MinkUNet from the available configurations in the repository (see \url{https://github.com/Pointcept/Pointcept/tree/main/configs/nuscenes}). In CleverDistiller, we introduce two additional components, namely the MLP projection head and the occupancy prediction task. For the occupancy prediction task, we use the same hyperparameters as used in ALSO \cite{also} (see \url{https://github.com/valeoai/ALSO}). The MLP projection head consists of $3$ layers with a hidden dimension of $2048$, based on the ablation experiment shown in \cref{table:ablation_mlp}. 

\paragraph{Choice of loss objective:} Following the motivation of \scalr \cite{scalr}, we use a cosine similarity based loss objective on the matching point and pixel features. Note that the point and pixel features are generally $\ell_2$-normalized before the loss. For such features, the MSE loss is equivalent to the cosine similarity loss. Moreover, the MSE formulation is more stable and used in \scalr as well. \scalr already compared the contrastive loss and the cosine similarity loss and found that the cosine similarity loss generally performed on par or better, while also being memory and compute efficient. This is because the cosine similarity loss requires computing a large similarity matrix of the contrasting features.

\begin{table}[ht]
  \label{table:app_hyperparams}
  \centering
  \footnotesize
  \begin{tabular}{lll}
    \toprule
    Hyper-parameter & MinkUNet & PointTransformerV3  \\
    \midrule
    batch size & $16$ & $16$ \\
    optimizer & AdamW & AdamW \\
    learning rate & $0.001$ & $0.002$ \\
    training epochs & $50 / 100$ & $50 / 100$ \\
    weight decay & $0.005$ & $0.005$ \\
    lr scheduler & OneCycleLR & OneCycleLR \\
    lr anneal strategy & cosine & cosine \\
    \midrule
    pointcloud augmentations & \multicolumn{2}{c}{RandomRotate, RandomScale, RandomFlip, RandomJitter} \\
    Mix3D prob & \multicolumn{2}{c}{$0.8$} \\
    coordinates & \multicolumn{2}{c}{cartesian} \\
    grid size & \multicolumn{2}{c}{$0.05$} \\
    \bottomrule
  \end{tabular}
  \vspace{0.2cm}
  \caption{Hyperparameter settings for ScaLR and CleverDistiller}
\end{table}

\section{Computational Cost}
\label{sec:comp_cost}
We run all the experiments on 4 x NVIDIA A100-80 GB GPUs. The CleverDistiller training using a MinkUNet-34 backbone over 100 epochs takes $\approx\!23\text{–}24$ hours depending on the size of the vision model (ViT-Small/Base/Large). Using a PointTransformer-V3 backbone, the same training takes around $\approx\!22.5\text{–}24.5$ hrs. Training a linear probe for 50 epochs on the full nuScenes dataset using frozen MinkUNet-34 and PointTransformer-V3 backbones take $\approx\!4$ and $\approx\!6$ hrs respectively. Finetuning for 100 epochs on the full nuScenes dataset using the MinkUNet-34 and PointTransformer-V3 backbones take $\approx\!10.5$ and $\approx\!11$ hrs respectively. 

\section{Fine-tuning setup}
\label{sec:evaluation_setup}

All semantic segmentation evaluations (linear probing and limited/full data fine-tuning) are performed using the same hyperparameter setups as described above using the configurations from the Pointcept framework for MinkUNet and PointTransformer-V3 backbones. Prior works tune the hyperparameters for each fine-tuning experiment -- for every dataset and scanario (LP, 1\%, 5\%, 10\%, 25\% and full data). 
We found that the performance achieved additionally using the Mix3D augmentation during the fine-tuning step generally matched the best performance achieved by tuning the hyperparameters. This removes the need for hyperparameter tuning specific to each experiment and makes the fine-tuning experiments easier and efficient. 

All the methods (our and SOTA) are pre-trained in a self-supervised way on the full nuScenes training set. For Linear Probing, the backbone stays frozen and only one linear layer is trained. For limited data fine-tuning experiments, we sampled the required amount of annotated \lidar frames randomly uniformly from all the available annotated \lidar frames:
\begin{itemize}
    \item 1,5,10 and 25 \% from $\approx$28,000 training frames in the nuScenes training split
    \item 1\% from $\approx$11,000 for KITTI training split
    \item 1\% from $\approx$24,000 for Waymo training split
\end{itemize}
Then, we use the full validation sets for the evaluation, similar to previous work \cite{superflow, slidr, xu2025limoe}. For other partial data splits used in the domain generalization experiments, we create the data splits following SuperFlow \cite{superflow}.

\section{Additional Experiments}
\label{sec:app_additional_expts}

\subsection{Mix3D in distillation training}
We use the Mix3D \cite{mix3d} augmentation also during pre-training and show an ablation study for this choice in \cref{table:ablation_mix3d}. We find that Mix3D provides a modest performance boost, when used during the knowledge distillation pre-training. We use the Mix3D augmentation for pre-training both \scalr and our CleverDistiller.

\begin{table}[h!]
\centering
\begin{tabular}{lcccc}
\toprule
\multirow{2}{*}{Mix3D} & \multicolumn{2}{c}{nuScenes} & KITTI & Waymo \\
\cmidrule(lr){2-3} \cmidrule(lr){4-4} \cmidrule(lr){5-5}
& LP & 1\% & 1\% & 1\% \\
\midrule
\textcolor{red}{\xmark} & 48.95 & 55.36 & 49.03 & 50.09 \\
\cellcolor{white!60!lightgray}\textcolor{green}{\cmark} & \cellcolor{white!60!lightgray}\textbf{49.81} & \cellcolor{white!60!lightgray}\textbf{56.90} & \cellcolor{white!60!lightgray}\textbf{50.59} & \cellcolor{white!60!lightgray}\textbf{50.99} \\
\bottomrule
\end{tabular}
\vspace{.2cm}
\caption{Ablation study of Mix3D during distillation}
\label{table:ablation_mix3d}
\end{table}

\subsection{3D Object Detection}
\label{sec:3dod}

For 3DOD, we use the same features distilled in SSL as we do for segmentation. This means we have one model, one distillation run, and the same features for multiple tasks, ensuring efficiency and consistency. 
3D backbone is pretrained in an unsupervised manner and then used as a backbone in the 3DOD task. Unlike other methods such as Olivine~\cite{olivine} and CMCR~\cite{zhang2024contrastive} (weakly supervised distillation methods which also report 3DOD), which need to distill features to a different network architecture like VoxelNet, our approach uses MinkUNet, making the features useful for both segmentation and 3DOD.

In our experiments shown in \cref{table:3dod}, we benchmarked our method on the KITTI and Waymo datasets.Our method consistently outperformed the previous best method by approximately 5-10\%, showing that the features are versatile and effective across multiple tasks.

We do not include results on the nuScenes dataset, as point-based methods are rarely evaluated on it due to the sparsity and low density of its point clouds, which limits their effectiveness. Additionally, most existing cross-modal distillation approaches report results only on KITTI and Waymo, making nuScenes unsuitable for fair and consistent comparison.
\begin{table}[h]
\renewcommand{\arraystretch}{0.5} 
\setlength{\tabcolsep}{10pt}
\scriptsize
\centering
\begin{tabular}{lccccc}
\toprule
\multirow{2}{*}{Method} & \multicolumn{3}{c}{KITTI} & \multicolumn{2}{c}{Waymo} \\
\cmidrule(lr){2-4} \cmidrule(lr){5-6}
 & 5\% & 10\% & 20\% & 5\% & 10\% \\
\midrule
Random init. & 56.1 & 59.1 & 61.6 & 18.0 & 21.9 \\
PPKT$^{50}$ & 57.8 & 60.1 & 61.2 & - & - \\
SLidR$^{50}$ & 57.8 & 61.4 & 62.4 & - & - \\
\textbf{CleverDistiller$^{50}$} & \cellcolor{white!70!SpringGreen}\textbf{58.3} & \cellcolor{white!70!SpringGreen}\textbf{63.7} & \cellcolor{white!70!SpringGreen}\textbf{66.8} & \cellcolor{white!70!SpringGreen}\textbf{22.7} & \cellcolor{white!70!SpringGreen}\textbf{24.6} \\
\cmidrule(lr){2-6}
ScaLR$^{100}$ & 56.2 & 62.3 & 66.5 & 20.1 & 28.1 \\
\textbf{CleverDistiller$^{100}$} & \cellcolor{white!10!SpringGreen}\textbf{59.8} & \cellcolor{white!10!SpringGreen}\textbf{66.6} & \cellcolor{white!10!SpringGreen}\textbf{67.1} & \cellcolor{white!10!SpringGreen}\textbf{22.2} & \cellcolor{white!10!SpringGreen}\textbf{28.7} \\
\bottomrule
\end{tabular}
\vspace{0.2cm}
\caption{\small Evaluation of our pre-training approach on 3D object detection benchmarks. All results in mAP.}
\label{table:3dod}
\end{table}

\section{Class-wise results}
\label{sec: app_classwise_results}

In \cref{tab:classwise} we compare the class-wise performance of CleverDistiller with other methods after finetuning on 1\% of the training data. We generally outperform other methods, including the most recent LiMoE on most classes \cref{tab:classwise}.


\begin{table*}[h!]
\footnotesize
\renewcommand{\arraystretch}{0.5} 
\setlength{\tabcolsep}{2.0pt}
\centering
\begin{tabular}{lccccccccccccccccc}
\toprule
\textbf{Method} & \rotatebox{90}{\textbf{Overall}} & \rotatebox{90}{\textbf{barrier}} & \rotatebox{90}{\textbf{bicycle}} & \rotatebox{90}{\textbf{bus}} & \rotatebox{90}{\textbf{car}} & \rotatebox{90}{\textbf{construction vehicle}} & \rotatebox{90}{\textbf{motorcycle}} & \rotatebox{90}{\textbf{pedestrian}} & \rotatebox{90}{\textbf{traffic cone}} & \rotatebox{90}{\textbf{trailer}} & \rotatebox{90}{\textbf{truck}} & \rotatebox{90}{\textbf{driveable surface}} & \rotatebox{90}{\textbf{other flat}} & \rotatebox{90}{\textbf{sidewalk}} & \rotatebox{90}{\textbf{terrain}} & \rotatebox{90}{\textbf{manmade}} & \rotatebox{90}{\textbf{vegetation}} \\
\midrule
\multicolumn{18}{c}{\cellcolor{white!60!lightgray}\textbf{ViT-S}} \\

SLidR$^{50}$ & 41.2 & 0.0 & 0.0 & 26.6 & 72.0 & 12.4 & 15.8 & 51.4 & 22.9 & 11.7 & 35.3 & 92.9 & 36.3 & 58.7 & 63.6 & 81.2 & 82.3 \\
Seal$^{50}$  & 44.3 & 20.0 & 0.0 & 19.4 & 74.7 & 10.6 & 45.7 & 60.3 & 29.2 & 17.4 & 38.1 & 93.2 & 26.0 & 58.8 & 64.5 & 81.9 & 81.9 \\ 
\textbf{CleverDistiller$^{50}$} & 53.7 & 59.1 & 0.0 & 61.4 & 83.4 & 1.2 & 40.1 & 62.1 & 23.1 & 28.0 & 56.3 & 93.7 & 53.3 & 62.2 & 68.8 & 83.5 & 82.6 \\
\midrule
ScaLR$^{100}$ & 53.0 & 58.4 & 0.0 & 64.9 & 83.3 & 10.5 & 20.7 & 59.6 & 24.6 & 26.4 & 55.8 & 93.9 & 51.2 & 63.4 & 68.5 & 83.7 & 83.1 \\ 
SuperFlow$^{150}$ & 47.8 & 38.2 & 1.8 & 25.8 & 79.0 & 15.3 & 43.6 & 60.3 & 0.0 & 28.4 & 55.4 & 93.7 & 28.8 & 59.1 & 59.9 & 83.5 & 83.1 \\
LiMoE$^{150}$ & 49.6 & 39.9 & 4.6 & 27.3 & 80.2 & 17.1 & 45.4 & 61.2 & 6.2 & 29.5 & 58.4 & 94.0 & 34.2 & 62.3 & 64.6 & 84.1 & 84.5 \\ 
\textbf{CleverDistiller$^{100}$} &  56.9 & 60.2 & 0.0 & 71.2 & 84.9 & 11.8 & 47.1 & 65.1 & 26.6 & 31.4 & 61.7 & 94.2 & 54.1 & 64.9 & 69.5 & 84.2 & 83.5 \\

\midrule \midrule
\multicolumn{18}{c}{\cellcolor{white!60!lightgray}\textbf{ViT-B}} \\

PPKT$^{50}$  & 40.9 & 0.0 & 0.0 & 24.5 & 73.5 & 12.2 & 7.0 & 51.0 & 13.5 & 15.4 & 36.3 & 93.1 & 40.4 & 59.2 & 63.5 & 81.7 & 82.2 \\ 
SLidR$^{50}$  & 41.6 & 0.0 & 0.0 & 26.7 & 73.4 & 10.3 & 16.9 & 51.3 & 23.3 & 12.7 & 38.1 & 93.0 & 37.7 & 58.8 & 63.4 & 81.6 & 82.7 \\ 
Seal$^{50}$  & 46.0 & 43.0 & 0.0 & 26.7 & 81.3 & 9.9 & 41.3 & 56.2 & 0.0 & 21.7 & 51.6 & 93.6 & 42.3 & 62.8 & 64.7 & 82.6 & 82.7 \\ 
\textbf{CleverDistiler$^{50}$}  & 55.8 & 60.7 & 0.0 & 68.8 & 85.1 & 22.0 & 28.9 & 61.8 & 27.6 & 30.1 & 61.0 & 94.0 & 52.9 & 63.7 & 69.1 & 84.0 & 82.9\\ 
\midrule
SuperFlow$^{150}$ & 48.1 & 39.1 & 0.9 & 30.0 & 80.7 & 10.3 & 47.1 & 59.5 & 5.1 & 27.6 & 55.4 & 93.7 & 29.1 & 61.1 & 63.5 & 82.7 & 83.6 \\ 
ScaLR$^{100}$ & 55.8 & 60.8 & 0.0 & 69.7 & 85.1 & 20.4 & 30.6 & 60.8 & 27.8 & 29.5 & 58.1 & 94.2 & 55.6 & 64.3 & 69.3 & 84.1 & 83.1 \\ 
LiMoE$^{150}$ & 50.2 & 41.5 & 3.8 & 32.2 & 81.7 & 12.9 & 49.3 & 61.1 & 7.3 & 29.3 & 57.8 & 94.2 & 35.1 & 62.9 & 65.4 & 84.0 & 84.8 \\ 
\textbf{CleverDistiller$^{100}$} & 59.8 & 61.8 & 0.0 & 72.8 & 85.7 & 35.2 & 49.6 & 65.4 & 33.6 & 32.9 & 62.6 & 94.5 & 59.5 & 65.6 & 69.5 & 84.7 & 83.5\\ 

\midrule \midrule
\multicolumn{18}{c}{\cellcolor{white!60!lightgray}\textbf{ViT-L}} \\

PPKT$^{50}$  & 42.1 & 0.0 & 0.0 & 24.4 & 78.8 & 15.1 & 9.2 & 54.2 & 14.3 & 12.9 & 39.1 & 92.9 & 37.8 & 59.8 & 64.9 & 82.3 & 83.6 \\ 
SLidR$^{50}$  & 42.8 & 0.0 & 0.0 & 23.9 & 78.8 & 15.2 & 20.9 & 55.0 & 28.0 & 17.4 & 41.4 & 92.2 & 41.2 & 58.0 & 64.0 & 81.8 & 82.7 \\ 
Seal$^{50}$  & 46.3 & 41.8 & 0.0 & 23.8 & 81.4 & 17.7 & 46.3 & 58.6 & 0.0 & 23.4 & 54.7 & 93.8 & 41.4 & 62.5 & 65.0 & 83.8 & 83.8 \\ 
\textbf{CleverDistiller$^{50}$} & 56.7 & 63.4 & 0.0 & 69.7 & 85.4 & 19.3 & 29.4 & 62.1 & 30.4 & 31.9 & 62.4 & 94.3 & 56.2 & 65.3 & 70.0 & 84.1 & 83.0 \\ 
\midrule
SuperFlow$^{150}$ & 50.0 & 44.5 & 0.9 & 22.4 & 80.8 & 17.1 & 50.2 & 60.9 & 21.0 & 25.1 & 55.1 & 93.9 & 35.8 & 61.5 & 62.6 & 83.7 & 83.7 \\ 
LiMoE$^{150}$ & 51.4 & 45.3 & 4.1 & 25.3 & 82.2 & 18.4 & 52.5 & 61.8 & 22.3 & 26.4 & 56.2 & 94.3 & 37.6 & 63.3 & 63.9 & 84.4 & 85.0\\ 
ScaLR$^{100}$ & 55.8 & 65.0 & 0.0 & 68.7 & 85.3 & 15.9 & 27.7 & 61.4 & 28.3 & 31.0 & 61.8 & 94.2 & 51.6 & 64.6 & 70.0 & 84.1 & 83.0  \\ 
\textbf{CleverDistiller$^{100}$}& 60.6 & 65.5 & 0.0 & 72.3 & 88.3 & 37.2 & 37.4 & 65.1 & 35.3 & 40.9 & 71.9 & 94.5 & 55.8 & 66.5 & 70.6 & 85.0 & 83.8\\ 

\bottomrule
\end{tabular}
\vspace{0.2cm}
\caption{Per-class IoU scores of state-of-the-art pretraining methods pretrained and fine-tuned on nuScenes with 1\% annotations. All IoU scores are given in percentage (\%).}
\label{tab:classwise}
\end{table*}


\section{Robustness - fine tuning}
\label{sec:robust_full}
The results in the table demonstrate that CleverDistiller achieves competitive and often superior performance compared to existing methods across a wide range of sensor corruptions. Notably, our method outperforms others in average and full mIoU for both 50 and 100 epochs labeled settings. At 100 epoch fine-tuning, the entire network is updated. This can potentially overwrite the structure learned during distillation, especially when the pretraining and fine-tuning objectives diverge. Similar performance saturation or degradation at full fine-tuning has been reported in earlier works. Nevertheless, CleverDistiller maintains the highest scores across most corruption types, validating the robustness and effectiveness of our distillation framework.
\begin{table*}[h!]
\footnotesize
\renewcommand{\arraystretch}{0.5} 
\setlength{\tabcolsep}{1.4pt}
\centering
\begin{tabular}{ccccccccccccccc}
\toprule
 & \multirow{2}{*}{Method} & \multirow{2}{*}{Backbone} & {\multirow{2}{*}{mCE $\downarrow$}} & {\multirow{2}{*}{mRR $\uparrow$}} & \multicolumn{9}{c}{mIoU $\uparrow$} \\
\cmidrule(lr){6-14}
 & \multicolumn{2}{l}{} & {} & {} & {Fog}   & {Rain} & {Snow}  & {Blur} & {Beam} & {Cross} & {Echo} & {Sensor} & Avg Full \\ 
 \midrule
 \multirow{18}{*}{Full} & {Random} & MinkU-34 & 112.20 & 72.57 & {62.96} & {70.65} & {55.48} & {51.71} & {62.01} & {31.56} & {59.64} & {39.41} & 54.18 \\ 

 \arrayrulecolor{lightgray} \cmidrule(lr){2-14}
 
 & {PPKT$^{50}$~\cite{ppkt}}  & MinkU-34 & 105.64 & 75.87 & {64.01} & {72.18} & {59.08} & {57.17} & {63.88} & {36.34} & {60.59} & {39.57} & 56.60 \\

 \cmidrule(lr){2-14}
 
 & {SLidR$^{50}$~\cite{slidr}} & MinkU-34 & 106.08 & 75.99 & {65.41} & {72.31} & {56.01} & {56.07} & {62.87} & {41.94} & {61.16} & {38.90}  & 56.83 \\

 \cmidrule(lr){2-14}
 
 & {Seal$^{50}$~\cite{seal}}  & MinkU-34 & \cellcolor{white!80!ForestGreen}\textbf{92.63} & 83.08 & {72.66} & \cellcolor{white!80!ForestGreen}{\textbf{74.31}} & 66.22 & \cellcolor{white!80!ForestGreen}{\textbf{66.14}} & \cellcolor{white!80!ForestGreen}{\textbf{65.96}} & {57.44} & {59.87} & {39.85}  & 62.81 \\

 \cmidrule(lr){2-14}
 
 & {\textbf{CleverDistiller}$^{50}$} & MinkU-34 &94.90 & \cellcolor{white!80!ForestGreen}\textbf{84.12}& \cellcolor{white!80!ForestGreen}{\textbf{72.73}} & {70.91} & \cellcolor{white!80!ForestGreen}{\textbf{67.35}} & {56.42} & {64.14} & \cellcolor{white!80!ForestGreen}{\textbf{63.76}} & \cellcolor{white!80!ForestGreen}{\textbf{60.17}} & \cellcolor{white!80!ForestGreen}{\textbf{47.27}} & \cellcolor{white!80!ForestGreen}\textbf{63.21} \\ 
 \cmidrule[2pt](lr){2-14}

 & {SuperFlow$^{150}$~\cite{superflow}} & MinkU-34 & 91.67 & 83.17 & {70.32} & {75.77} & {65.41} & {61.05} & {68.09} & {60.02} & {58.36} & {50.41}  & 63.68 \\

 \cmidrule(lr){2-14}
 
 & {Scalr$^{100}$~\cite{scalr}} & MinkU-34 & 99.86& 81.35& {69.23} & {72.47} & {60.67} & {55.11} & {64.07} & {56.93} & {60.34} & {46.11} & 60.62\\

 \cmidrule(lr){2-14}
                      
&{LiMoE$^{150}$} & MinkU-34 &\cellcolor{white!60!ForestGreen}\textbf{ 88.43} & 83.28 & {71.10} & \cellcolor{white!60!ForestGreen}{\textbf{75.92}} & {65.66} & \cellcolor{white!60!ForestGreen}{\textbf{63.86}} & \cellcolor{white!60!ForestGreen}{\textbf{68.52}} & {60.78} & \cellcolor{white!60!ForestGreen}{\textbf{61.91}} & {50.66} & 64.80 \\ 
\cmidrule(lr){2-14}
& {\textbf{CleverDistiller}$^{100}$} & MinkU-34 & 91.28 & \cellcolor{white!60!ForestGreen}\textbf{87.42} & \cellcolor{white!60!ForestGreen}{\textbf{72.83}} & {73.33} & \cellcolor{white!60!ForestGreen}{\textbf{66.03}} & {60.18} & {66.85} & \cellcolor{white!60!ForestGreen}{\textbf{68.04}} & {60.65} & \cellcolor{white!60!ForestGreen}{\textbf{53.63}} & \cellcolor{white!60!ForestGreen}\textbf{64.83} \\ 

\arrayrulecolor{black} \midrule \midrule \arrayrulecolor{lightgray}

\arrayrulecolor{black} \bottomrule
\end{tabular}
\vspace{0.2cm}
\caption{3D robustness study of state-of-the-art pretraining methods under corruption and sensor failure scenarios in the nuScenes-C dataset from the Robo3D benchmark~\cite{kong2023robo3d}. Full denotes fine-tuning with 100\% labeled data.  All mCE (↓), mRR (↑), and mIoU (↑) scores are given in percentage (\%). Best scores in each configuration are shaded with colors.}
\label{tab:app_robust}
\end{table*}

\section{Temporal consistency}
\label{sec:temporal}

\begin{table*}[h!]
\footnotesize
\renewcommand{\arraystretch}{1.2} 
\setlength{\tabcolsep}{1.4pt}
\centering
\begin{tabular}{ccc}
\toprule
\textbf{Temporal} & \textbf{Occupancy} & \textbf{LP (mIoU)} \\
\midrule
\textcolor{red}{\xmark} & \textcolor{red}{\xmark} & 40.76 \\
\textcolor{green}{\cmark} & \textcolor{red}{\xmark} & 41.38 \\
\textcolor{red}{\xmark} & \textcolor{green}{\cmark} & \textbf{ 49.81} \\
\textcolor{green}{\cmark} & \textcolor{green}{\cmark} & 44.53 \\
\bottomrule
\end{tabular}
\vspace{0.2cm}
\caption{Ablation of Minkunet pre-trained on nuScenes with our method and then evaluated with Linear Probing (LP) on semantic segmentation task on nuScenes. We test different auxilary tasks: Temporal Consistency and Occupancy prediction. We report mIoU.}
\label{tab:temporal}
\end{table*}

To identify temporally consistent regions, we first remove the ground plane from each frame. Then, we project two point clouds from temporally adjacent frames into a shared global coordinate frame. Points that spatially overlap across these frames are considered temporally consistent. We posit that features corresponding to these overlapping regions should also be consistent in the learned feature space. This motivates our temporal consistency loss, which encourages the model to produce stable and coherent features over time.

However, in practice, we observe that this temporal constraint is less effective than the occupancy prediction objective (see \cref{tab:temporal}). While temporal consistency captures some structural regularities, it does not lead to stronger downstream performance on its own. Moreover, when combined with the occupancy prediction task, it slightly degrades performance compared to using occupancy prediction alone. This suggests that the temporal consistency loss may interfere with the spatial cues emphasized by occupancy supervision, indicating that it is not a complementary signal in our setting.

\section{Qualitative results}
\label{sec:visualization}

\subsection{Segmentation errors}

We can also observe in \cref{fig:combined} that our predictions have fewer errors along the segmentation boundaries than those from \scalr. Types of segmentation errors are not reflected by mIoU scores, but well-defined boundaries are crucial for the output quality. 
\begin{figure*}[h]
    \centering
    \includegraphics[trim=0 120 350 10, clip, width=\linewidth]{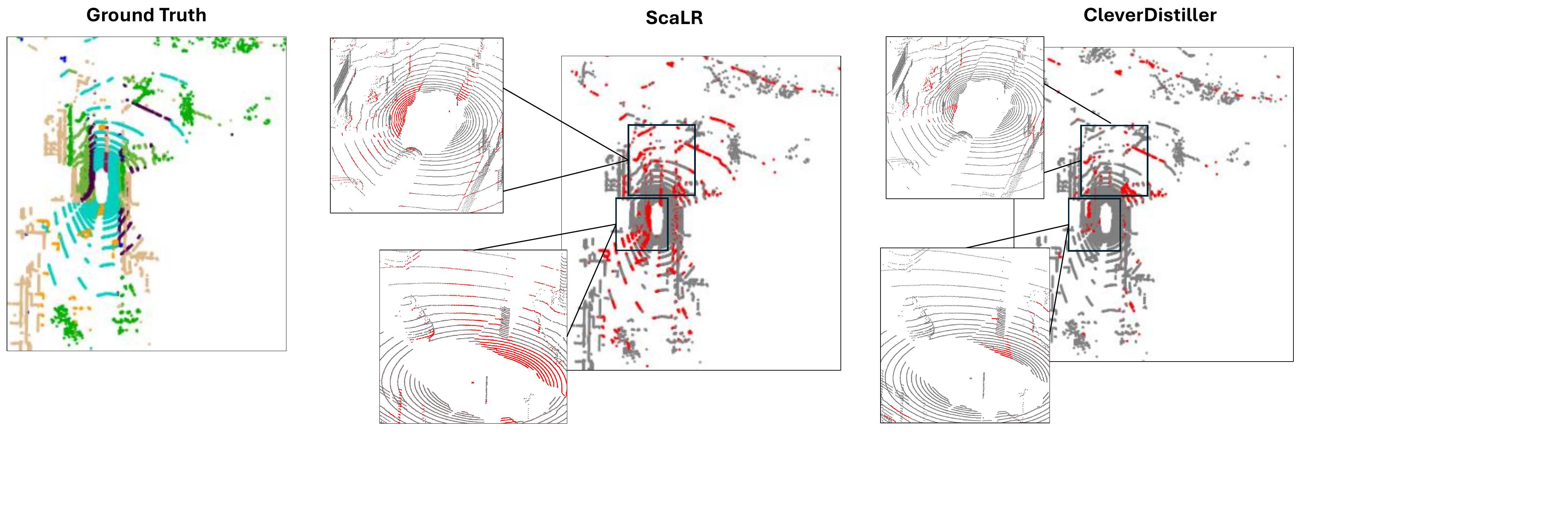}
    \label{fig:subfig2}
    \vspace{-.8cm}
\end{figure*}
\begin{figure*}[h]
    
        \centering
        \includegraphics[trim=0 275 530 10, clip, width=\linewidth]{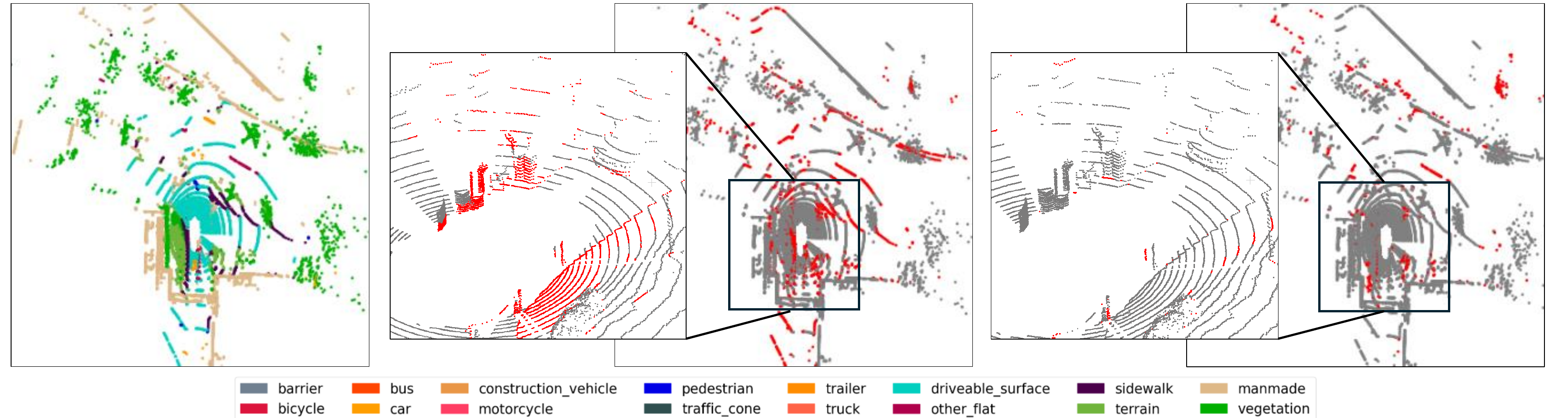}
    \vspace{.2cm}
    \caption{\small Qualitative segmentation results comparing our CleverDistiller (right) to ScaLR (middle) and the ground truth (left). Legend corresponds to the ground truth. For ScaLR~\cite{scalr} and CleverDistiller we show prediction errors (\textcolor{red}{in red}), showcasing that our approach exhibits much less errors. Each row is a different sequence randomly sampled from nuScenes~\cite{caesar2020nuscenes}.}
    \label{fig:combined}
\end{figure*}
\subsection{Qualitative features analysis}
\label{sec: app_qualitative_analysis}
\begin{figure*}[h]
    \centering
    \includegraphics[width=\linewidth]{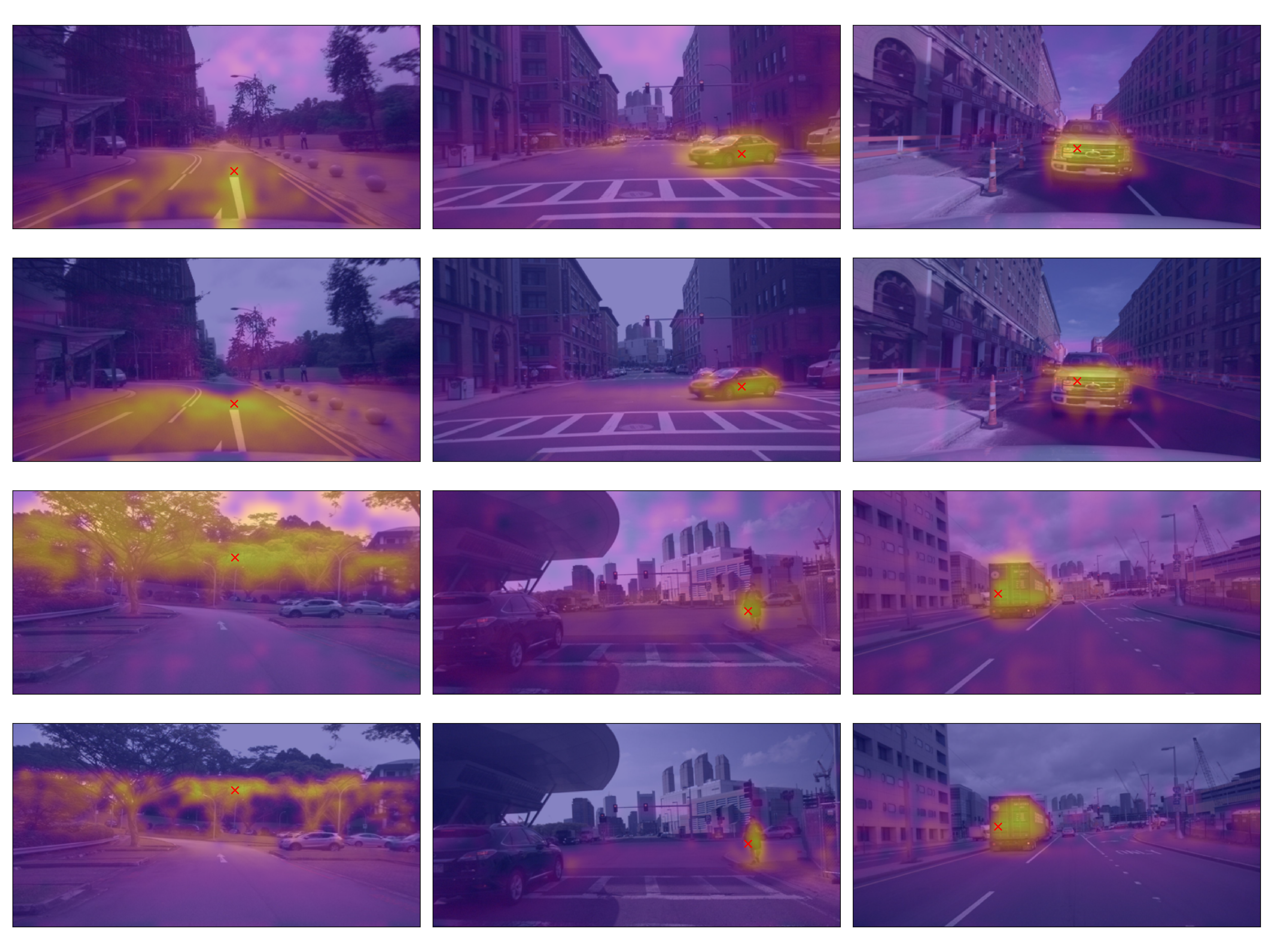}
    \caption{Visualization of feature similarities of CleverDistiller. Rows 1 and 3 show similarities of query pixel to other pixels based on teacher model's image features. Rows 2 and 4 show similarities of query point to other points based on the distilled backbone's features. Best viewed in color. The query points / pixels are denoted using red crosses. Yellow denotes high similarity and blue denotes low similarity.}
    \label{fig:feature_similarity_visualization_clever}
\end{figure*}

Similarity maps presented in \cref{fig:feature_similarity_visualization_clever} illustrate the segmentation ability of our pretrained model, CleverDistiller. The query points include “truck”, “car”, "driveable surface", and "vegetation". CleverDistiller shows strong semantic discriminative ability without fine-tuning, similar to features distilled from camera foundational models. This effectiveness can be attributed to three key aspects: 1) Cosine similarity loss, which ensures that features are aligned; 2) An MLP projection head that retains information during projections, unlike previous methods using linear layers, helping the features to become more aligned; 3) An additional 3D auxiliary spatial task that improves the 3D features and understanding of the 3D backbone. The visualization of feature similarities in Fig. 5 highlights how CleverDistiller's features align with those of the teacher model. Rows 1 and 3 show similarities of query pixels to other pixels based on the teacher model’s image features, while rows 2 and 4 show similarities of query points to other points based on CleverDistiller’s distilled backbone features. Yellow denotes high similarity and blue denotes low similarity, with query points/pixels marked by red crosses. Best viewed in color.

\end{document}